\documentclass[10pt,twocolumn,letterpaper]{article}

\usepackage[pagenumbers]{cvpr} 
\usepackage{multirow}
\usepackage{float}
\usepackage{amsmath}
\usepackage{dashrule} 
\usepackage{multirow}
\usepackage{amsmath}
\usepackage{booktabs}
\usepackage{graphicx}
\usepackage{amsfonts}
\usepackage{algorithm2e}
\usepackage{lipsum}
\usepackage{graphicx}
\usepackage{tabularx}
\usepackage{array}
\usepackage{caption}
\usepackage{dashrule} 

\usepackage{url}
\definecolor{cvprblue}{rgb}{0.21,0.49,0.74}
\usepackage[pagebackref,breaklinks,colorlinks,allcolors=cvprblue]{hyperref}

\def\paperID{2} 
\def\confName{CVPR}
\def\confYear{2025}

\title{FiLA-Video: Spatio-Temporal Compression for Fine-Grained Long Video Understanding}

\author{%
{Yanan Guo}\textsuperscript{\textnormal{1}},
{Wenhui Dong}\textsuperscript{\textnormal{2}}, 
{Jun Song}\textsuperscript{\textnormal{2}}, 
{Shiding Zhu}\textsuperscript{\textnormal{3}},
{Xuan Zhang}\textsuperscript{\textnormal{2}},
{Hanqing Yang}\textsuperscript{\textnormal{2}},\\
{Yingbo Wang}\textsuperscript{\textnormal{2}},
{Yang Du}\textsuperscript{\textnormal{2}},
{Xianing Chen}\textsuperscript{\textnormal{2}},
{Bo Zheng}\textsuperscript{\textnormal{2}}
\\ 
\textsuperscript{1}University of Science and Technology of China,  \textsuperscript{2}Alibaba Group, 
\textsuperscript{3}ZheJiang University\\
}

\begin{document}
\maketitle
\begin{abstract}
Recent advancements in video understanding within visual large language models (VLLMs) have led to notable progress. However, the complexity of video data and contextual processing limitations still hinder long-video comprehension. A common approach is video feature compression to reduce token input to large language models, yet many methods either fail to prioritize essential features, leading to redundant inter-frame information, or introduce computationally expensive modules.To address these issues, we propose \textbf{FiLA}(\textbf{Fi}ne-grained Vision \textbf{La}nguage Model)\textbf{-Video}, a novel framework that leverages a lightweight dynamic-weight multi-frame fusion strategy, which adaptively integrates multiple frames into a single representation while preserving key video information and reducing computational costs. To enhance frame selection for fusion, we introduce a keyframe selection strategy, effectively identifying informative frames from a larger pool for improved summarization. Additionally, we present a simple yet effective long-video training data generation strategy, boosting model performance without extensive manual annotation. Experimental results demonstrate that \textbf{FiLA-Video} achieves superior efficiency and accuracy in long-video comprehension compared to existing methods.
\end{abstract}    
 \begin{figure}
  \centering
  \includegraphics[width=.5\textwidth]{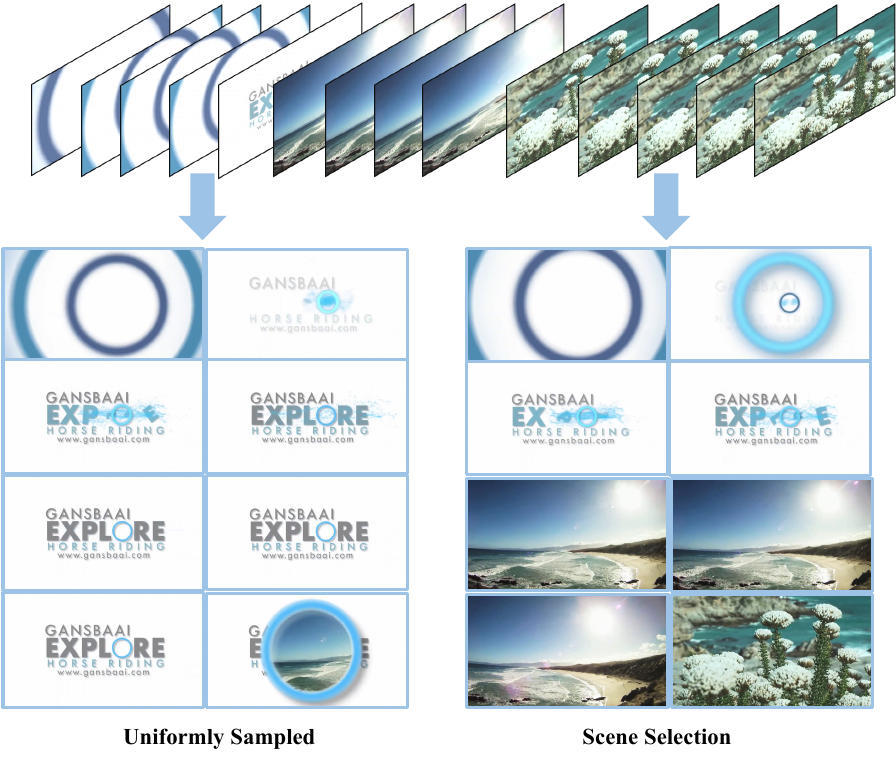} 
  \caption{Comparison between the frames selected by Uniformly Sampling and Scene-Selection respectively. We show the first 8 frames collected from the same video. The left pair shows that the method of uniformly sampling frames inevitably collects too many repeated frames, which leads to a waste of computing resources during token compression. The pair on the right solves the problem.}
  \label{fig:image_scene} 
\end{figure}
\section{Introduction}
In recent years, multimodal large language models (MLLMs) such as \cite{mllm1llava,mllm2alayrac2022flamingovisuallanguagemodel,mllm3bai2023qwenvlversatilevisionlanguagemodel,mllm4chen2024internvlscalingvisionfoundation,mllm5liu2024improvedbaselinesvisualinstruction,mllm6zhu2023minigpt4enhancingvisionlanguageunderstanding} have made significant progress in the field of visual understanding. Due to the progress of these multimodal large language models in image understanding, the field of video understanding can well transfer their methods. Studies such as \cite{qwen2-vl}, \cite{llavaonevision}, \cite{llava-video} have demonstrated the effectiveness of extending multimodal text-image large models to the video domain. However, due to the characteristics of video data that are composed of multiple image frames, how to handle an excessively long number of visual tokens has become a challenge. The existing multi-modal information processing paradigm mainly relies on LLM to process the encoded features. Therefore, the context limitations of LLM and the memory limitations restrict the processing of long videos. To address the issue, a considerable amount of research has focused on compressing video features to reduce the number of tokens ultimately submitted to the LLM. For example, VideoLLaMA2 \cite{videollama2} adopts the STC connector to implement spatial-temporal aggregation, preserving spatial and temporal
local details while reducing the number of visual tokens.  Video-ChatGPT \cite{videochatgpt} performs average-pooling on visual embeddings along spatial and temporal dimensions, respectively. LLaMA-Vid \cite{llamavid} utilizes text instruction to generate a contextual token and performs down-sampling on the visual embeddings of a long video as another token.

However, current mainstream methods have several drawbacks. Firstly, these methods often begin by uniformly sampling a small portion of frames for processing \cite{llavaonevision, llava-video, nvila}, which might overlook important frames while capturing less critical ones, leading to inadequate video understanding. Secondly, there is still room for optimization in feature processing. For example, some methods \cite{timechat, self-chained} introduce a module for feature merging that is overly complex, leading to a more complicated model structure, increased computation, and reduced transferability. Simple down-sampling methods \cite{tavg}, commonly used in works like \cite{videochatgpt, fromimagetovideo}, lack learnable characteristics. Current merging methods infrequently consider the entire video when determining which frames to merge, often damaging the event and scene information within videos, as some frames are unsuitable for merging together. Thirdly, current video understanding datasets still lack long video datasets, which is due to the lack of high-quality long-video data sources on the Internet. For example, ShareGPT4Video \cite{sharegpt4video} mainly consists of 30-second videos, and llava-video-178k \cite{llava-video} mainly consists of videos under 3 minutes.

To address these issues, we propose FiLA-Video, a novel framework for video understanding, which introduces an efficient method for spatio-temporal feature compression based on the selection of important scenes. We represent each frame with the mean of its patch tokens and use a clustering algorithm to identify the important scenes, which typically consist of a large number of similar frames. At the same time, we can obtain clustering centers as representative frames of important scenes and then select additional frames to form different scenes strictly in chronological order. Meanwhile, this approach also avoids the problem of clustering dissimilar frames together. For each scene, we focus only on temporal features and introduce a simple yet effective merging method by applying learnable weights to each patch token, greatly reducing computational complexity and simplifying the model structure, while demonstrating superior performance compared to uniform frame sampling, cross-attention, and spatio-temporal down-sampling methods.

Moreover, we introduce a pipeline that utilizes high-quality short-video datasets to generate long-video captioning datasets. This method does not rely on public long videos on the Internet or complex annotation methods, yet it can still improve the model's ability to understand long videos. Specifically, we constructed a visual captioning dataset by combining multiple short videos and their annotations. To enable the model to more accurately understand the temporal relationships within videos, we enrich the dataset with temporal cues, where video captions are segmented by scenes and correspond to specific timestamps. Experimental results show that even with the simplest uniform frame sampling method, the model can learn more knowledge from this part of the data.

Overall, our contributions are as follows:
\begin{itemize}
    \item \textit{Vision-Language Model for Video Understanding}: We propose FiLA-Video, a flexible framework which combines a improved clustering method to select important scenes in long videos and a simple and effective method to merge features within the scene in the spatio-temporal dimensions. Through experiments, we demonstrate its superiority over other methods.
    \item \textit{Long Video Captioning Dataset}: We introduce a long-video captioning dataset enriched with extensive timestamp information, where all data range from 5 to 30 minutes in duration. This dataset shows significant improvements on popular benchmarks with small samples.
    \item \textit{Open-Source}: We release our model checkpoints, code and dataset to the public, aiming to promote the progress of the open-source community.
\end{itemize}
\section{Related Work}
\label{sec:formatting}
\subsection{MLLM for Video Understanding}
The remarkable performance of multimodal large language models (MLLMs) has significantly advanced the field of visual understanding by effectively integrating textual and visual information. These models, such as Flamingo \cite{mllm2alayrac2022flamingovisuallanguagemodel}, LLaVA \cite{liu2023visualinstructiontuning} , extend the capabilities of traditional language models by incorporating visual context through sophisticated fusion techniques. Flamingo integrates image features with language models via a gated cross-attention mechanism, allowing the model to dynamically weigh visual and textual inputs for enhanced comprehension tasks. Approaches like LLaVA and MiniGPT4 \cite{zhu2023minigpt4enhancingvisionlanguageunderstanding} simplify the process by directly projecting visual features into the LLM embedding space through a lightweight multi-layer perceptron (MLP), bypassing complex cross-modal fusion modules.

Thanks to the advancement of multi-modal large models, they can be easily transferred to tasks in the field of video understanding.
Models such as LLaVA-OneVision \cite{llavaonevision} and LLaVA-Next-Video \cite{zhang2024llavanextvideo}  employ similar unified frameworks to handle multiple modalities, which enables tasks such as video captioning and question answering. These models are often trained on image tasks, then are trained on video data. They treat the video as multiple frames to process, relying on pre-trained vision encoders, such as CLIP \cite{clip} or Siglip \cite{siglip} to extract relevant visual features, which are then aligned with textual data via contrastive learning. By exploiting large-scale datasets and self-supervised learning objectives, MLLMs like \cite{llavaonevision,llava-video,qwen2-vl,zhang2025videollama3frontiermultimodal} have demonstrated remarkable performance in various video understanding benchmarks, pushing the boundaries of what can be achieved in this domain.
\subsection{Video Feature Compression}
A mainstream approach in handling large numbers of visual tokens is to extract important information or merge them into a smaller number of tokens. A simple and effective method to achieve this is by applying pooling operations to the features, as seen in models like Video-ChatGPT \cite{videochatgpt}, Valley \cite{valley}, and NViLA \cite{nvila}. Another approach is binary soft matching, exemplified by ToMe \cite{tome}, LongVLM \cite{longvlm}, and Dyto \cite{dyto}. Some methods utilize learnable modules for extraction, such as Q-Former \cite{videollama,bvllm,auroracap,timechat} or convolution \cite{videollama2,ppllava} to compress visual features.

Another strategy is visual token pruning, which involves discarding unimportant visual tokens to reduce token consumption while minimizing performance loss. These methods focus on strategies for selecting which tokens to discard. Key techniques include FastV \cite{fastv}, Lyra \cite{lyra}, Feather the Throttle \cite{featherthrottle}, and DyCoke \cite{dycoke}.

Among methods for compressing visual features, some work not only focuses on merging visual tokens but also considers the relationships between tokens, as visual tokens from the same event or similar scenes should be prioritized for merging. For example, Chat-UniVi \cite{Chat-UniVi} and PruneVid \cite{prunevid} employ a KNN approach to cluster visual tokens, while Dyto uses the time-weighted distance of CLS tokens for clustering.
\subsection{Video Understanding Dataset}
There are currently many video datasets used to train or evaluate models' video understanding capabilities. Among popular benchmarks, VideoMME \cite{videomme} covers six major visual domains and 30 sub-domains, encompassing short, medium, and long videos, with video lengths ranging from 11 seconds to 1 hour. It includes 900 manually selected and annotated videos, totaling 254 hours, with 2,700 question-answer pairs generated. VideoBench \cite{videobench} selects approximately 15,000 QA pairs, covering ten evaluation dimensions. MVBench \cite{mvbench} challenges LVLMs with video tasks that cannot be effectively addressed by relying on single frames, comprising 4,000 QA pairs from 11 public video benchmarks. TempCompass \cite{tempcompass} focuses on assessing LVLMs' nuanced performance in various temporal aspects, such as speed, direction, and attribute changes, with 410 videos and 7,540 carefully collected instructions, emphasizing temporal understanding and interaction. Benchmarks specifically designed for long videos, such as Hourvideo \cite{hourvideo}, include 500 handpicked egocentric videos from the Ego4D \cite{ego4} dataset, ranging from 20 to 120 minutes in length, and include 12,976 high-quality multiple-choice questions with five options each. Cinepile \cite{cinepile} gathers a large collection of movie clips, covering themes like temporal understanding, perceptual analysis, and complex reasoning, comprising approximately 305,000 question-answer pairs from 9,396 videos.\\Datasets designed to enhance MLLM performance in video understanding tasks, such as ShareGPT4Video \cite{sharegpt4video}, contain 40k high-quality videos spanning a wide range of categories. The generated subtitles include rich world knowledge, object attributes, camera movements, and detailed and accurate temporal descriptions of events. LLaVA-Video-178K \cite{llava-video} contains 178,510 videos ranging from 0 to 3 minutes in length, sourced from ten datasets, covering a broad spectrum of video domains.

\section{Method}

In this section, we will provide a detailed introduction to FiLA-Video. Within this framework, we propose an efficient method for spatio-temporal feature compression based on the selection of important scenes. We hypothesize that important scenes in a video often form visual clusters, and by identifying these cluster centers, we can quickly locate the corresponding scenes. We use the average of patch tokens as the representative feature for each frame, then employ a clustering algorithm to select representative frames of important scenes. Subsequently, a certain proportion of frames are selected in strict chronological order to supplement and form different scenes. As shown in Figure \ref{fig:image_scene}, by selectively capturing frames in this way, we can obtain more valuable image information than uniformly sampling frames. For each scene, we focus solely on the temporal dimension of the features and introduce a simple yet effective merging method, which involves applying learnable weights to each patch token.
\subsection{Scene Selection}
\label{sceneselec}
\subsubsection{Clustering-based Scene Selection }
First, for the input set of \(N\) frames \( X = \{I_1, I_2, \ldots, I_N\} \), each frame is independently encoded by a visual encoder to obtain a sequence of visual features \( V = \{V_1, V_2, \ldots\} \). These features are subsequently processed by the scene selection module.

Next, we calculate the representative feature for each visual feature. For every frame \( V_i \) in the video, which consists of \( L \) patch tokens each with dimension \( D \), we average the patch tokens of each frame to serve as the representative feature. This converts the feature sequence from \( (N, L, D) \) to \( (N, D) \).
Specifically, for each image feature \( V_i \), represented as a matrix \( V_i = [p_1, p_2, \ldots, p_L] \) where \( p_j \) is the \(j\)-th patch token with dimension \( D \), the representative feature is computed as:
\begin{equation}
\label{eq:mean}
\text{P}_i = \frac{1}{L} \sum_{j=1}^{L} p_j 
\end{equation}

The calculated \( P_i \) serves as the representative feature for a frame, denoted as \( (N, D) \).

After caculating the representative features, we use the K-Means algorithm \cite{kmeans} to select representative frames of important scenes. The selection process is explained below:

\textbf{Initialization} Choose \( M \) random representative features as the initial cluster centers.

  \textbf{Assignment step} For each representative feature \( P_i \), calculate its Euclidean distance to each cluster center \( \mu_j \) and assign \( P_i \) to the nearest cluster center:
  \begin{equation}
  \operatorname{assign}(P_i) = \arg\min_{1 \leq j \leq M} \|P_i - \mu_j\|^2
  \end{equation}
  
  \textbf{Update step} For each cluster center \( \mu_j \), compute the mean of all its assigned features and update its position:

  \begin{equation}
  \mu_j = \frac{1}{|C_j|} \sum_{P_i \in C_j} P_i
  \end{equation}

  where \( C_j \) is the set of samples assigned to the j-th cluster center, and \(|C_j|\) is its count.

Repeat the iteration of assignment and update steps until the cluster centers converge or reach the maximum number of iterations. Eventually, the centers of these clusters will serve as the representative frames of the selected scenes. Then we get the index of those frames:
\begin{equation}
K = \{ \mu_1, \mu_2, \ldots, \mu_k \}
\end{equation}

Where \(k\) is the number of the selected scenes. \(\mu_i\) is the index of representative frames of corresponding scene.

We define that the historical frames are those frames indexed from \(\mu_{i-1}\) to \(\mu_i\). Based on this, the similarity scores between the representative frames and its historical frames are computed, and the most similar \(r\) frames are added to supplement the scene. For a given representative frame at index \(\mu_i\). The similarity scores are computed as follows:
\begin{equation}
\resizebox{0.5\hsize}{!}{$
\operatorname{sim}(P_{\mu_i}, P_j) = \frac{P_{\mu_i} \cdot P_j}{\|P_{\mu_i}\| \times \|P_j\|}
$}
\end{equation}
where \(P_{\mu_i} \cdot P_j\) is the dot product of two feature vectors, and \(\|\cdot\|\) represents the magnitude of the vector. We then select the top \(r\) frames with the highest cosine similarity scores as supplements. This can be expressed as:

\begin{equation}
\begin{aligned}
S_{\mu_i} = & \{V_{\mu_i}\} + \{V_j \mid \operatorname{sim}(P_{\mu_i}, P_j) \text{ is among the top } r, \\
            & \quad \mu_{i-1} < j < \mu_i\}
\end{aligned}
\end{equation}

where \(S_{\mu_i}\) represents the feature maps of a selected scene. Finally, we obtain \(k\) scenes, each containing \(r+1 \) frames.

\subsection{Scene Mergeing}
In this section, we will introduce in detail the methods we adopt as the merging module of our model.
\label{method:merge}
This approach uses a learnable and simple structure to achieve good performance. We introduce a fine-grained and learnable weighting mechanism to perform weighted fusion of image features within scenes. The input features have a shape of \((k, N, D)\), where  \(k\) is the scene size, 
\(N\) is the number of patches, and \(D\) is the dimension of each feature. The output features are merged into a comprehensive feature with a shape of \(( N, D)\) .By assigning different weights to the spatial positions of each frame within the scene, we achieve fine-grained weighted fusion along the temporal dimension.

Given a set of feature maps $\{F_i\}_{i=1}^k$ with shape $(k, N, D)$, we introduce a learnable weight parameter matrix \(\mathbf{W} \in \mathbb{R}^{k \times N \times D}\).

When initializing \(\mathbf{W}\), we average all the patch tokens over the time span:

\begin{equation}
\mathbf{W}_{ij} =\frac{1}{k} \mathbf{1}_D, \quad \forall i \in \{1, \ldots, k\}, \quad \forall j \in \{1, \ldots, N\}.
\end{equation}

Next, to obtain the output feature map \(F_{\text{out}}\), we perform a weighted sum over all input feature maps:

\begin{equation}
F_{\text{out}} = \sum_{i=1}^k \left( F_i \odot \mathbf{W}_i \right)
\end{equation}

Here, \(\odot\) denotes element-wise multiplication, and \(\mathbf{W}_i\) represents the \(i\)-th row of matrix \(\mathbf{W}\).

This approach not only balances the importance of different frames along the temporal dimension but also allows the trainable parameters to automatically adjust each patch's contribution during the training process, offering more flexible and precise feature representation for long-term video analysis tasks.

\begin{figure}[htbp]
  \centering
  \includegraphics[width=.5\textwidth]{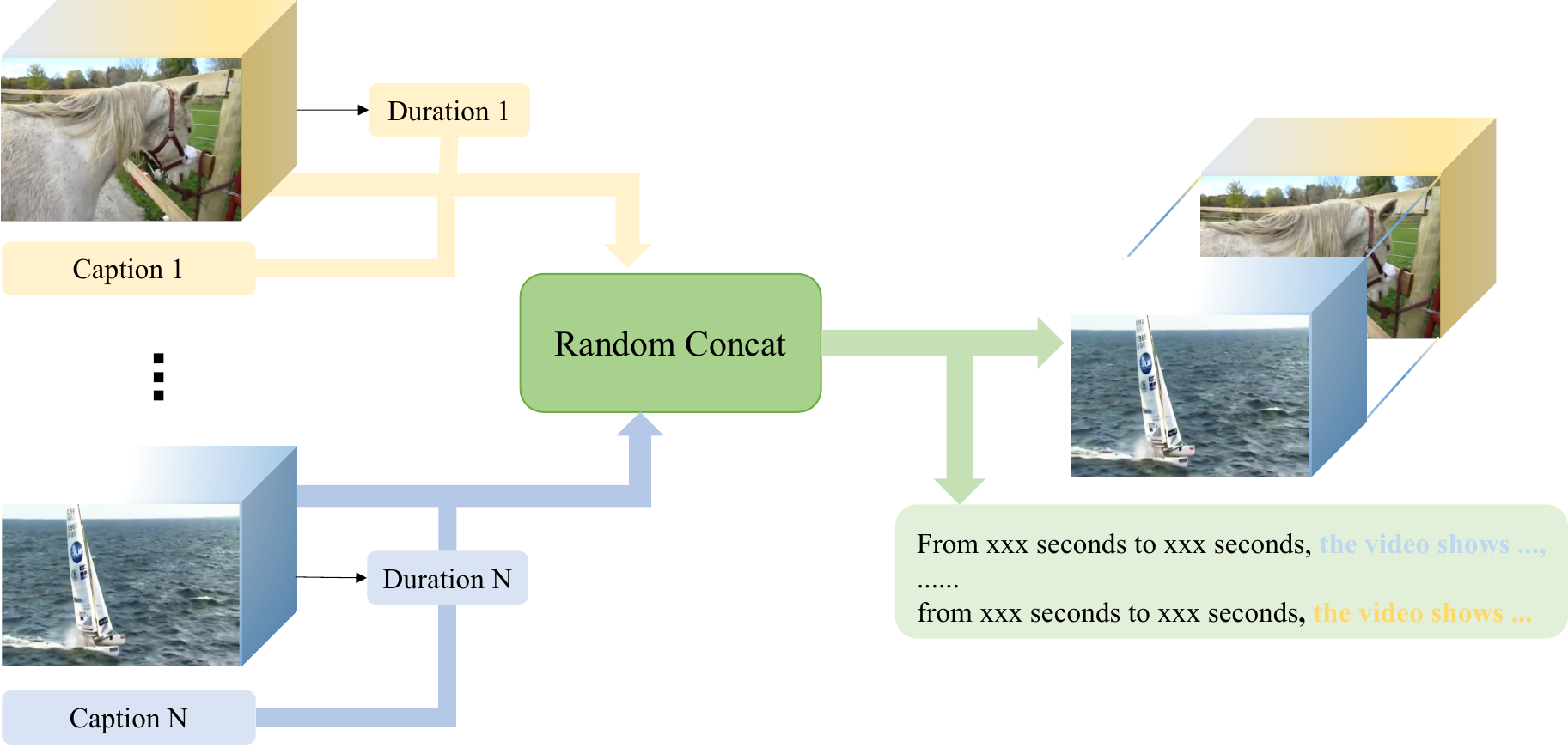} 
  \caption{The pipeline for our synthetic data.}
  \label{fig:figdata} 
\end{figure}
\subsection{LongVideo Dataset}
\label{method:data}
To address the current insufficiency of sufficiently long videos in video datasets, we have merged a large number of videos with visual description annotations into long videos of 5-30 minutes as shown in Figure \ref{fig:figdata}. We used the existing llava-video-178k data set \cite{llava-video} to synthesize this part of the data, and the experiments prove the effectiveness of the synthetic data. 

\begin{figure*}[t]
  \centering
  \includegraphics[width=.8\textwidth]{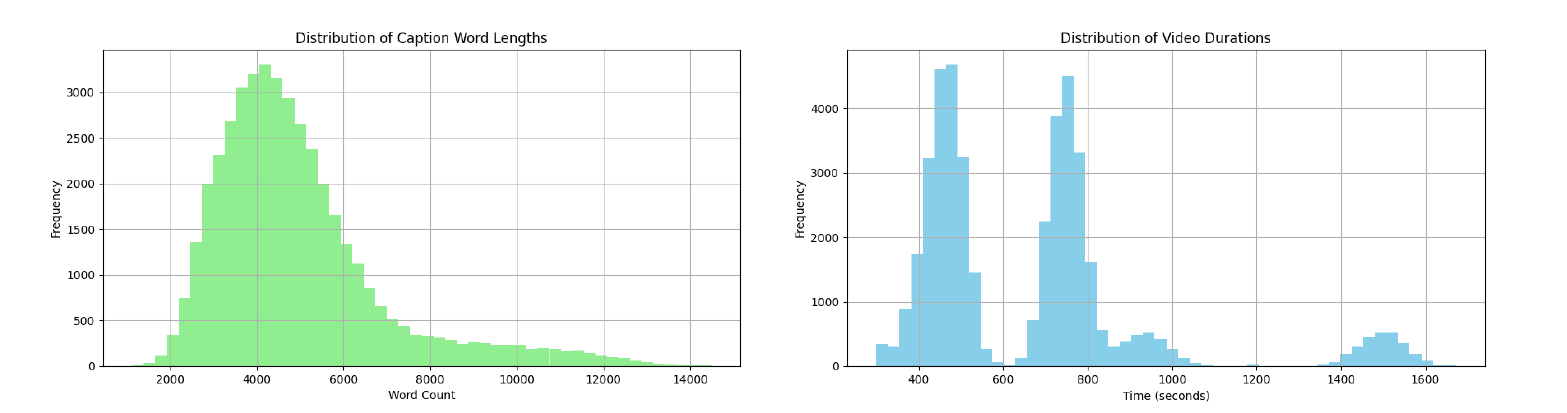} 
  \caption{(Left)Visualization of the video duration. (Right)Visualization of the video caption length.}
  \label{fig:stats} 
  \vspace{3mm}
\end{figure*}

Specifically, we consider each sub-video involved in the merging process as a scene in the final merged video. We obtain the dense subtitle timestamps for each scene based on the duration of the sub-videos and incorporate these timestamp informations into the final subtitles of the merged video. Additionally, similar to VideoChat \cite{videochat}, we incorporated the total duration of the video and the timestamp information of the sampled frames into the user instructions during training. For example, We insert `` This video samples \(N\) frames of a \(T\)-second video at \(t1, t2, ...\) seconds." to the instruction. Experiments show that even when mixed with a data-to-synthetic data ratio of approximately \(40:1\) and training with the simplest 32-frame uniform sampling, the model can also achieve significant performance improvements on benchmarks such as VideoMME \cite{videomme}. Figure \ref{fig:stats} shows the statistics of the synthetic data. We counted the distribution of caption size and video durations separately.

\subsection{Training Strategy}
We draw on the training strategy of LLaVA-OneVision \cite{llavaonevision} and adopt a four-stage training process. Our training is divided into four stages: vision-language alignment, knowledge injection, image-only instruction tuning, and video-only instruction tuning. This phased training strategy greatly aids in decoupling the experiments, allowing us to explore the impact of different data at various stages on the model's performance.

\textbf{Stage 1: Vision-Language Alignment}. In this stage, we use BLIP-558K. The objective of this stage is to align visual features with the text embedding space of the large language model (LLM).

\textbf{Stage 2: Knowledge Injection}. In this stage, we use 4.0 million samples from  open-sourced high-quality knowledge dataset \cite{llavaonevision}, which includes Document / OCR data and re-captioned detailed description data generated with  COCO118K, BLIP558K, and CC3M. The goal of this stage is to enable the multimodal LLM (MLLM) to learn high-quality knowledge such as mathematical concepts.

\textbf{Stage 3: Image-only Instruction Tuning}. In this stage, we perform instruction tuning using the single-image data from the OneVision dataset.

\textbf{Stage 4: Video-only Instruction Tuning}. In this stage, we employ LLaVA-Video-178K which contains 1.6 million samples for instruction tuning. Additionally, At this stage, we do not introduce the scene selection module for training. Instead, we first perform multi-rate uniform frame sampling, and then use the scene merging module to merge the features to the normal number of frames, similar to \cite{nvila}, specifically, we uniformly sample 96 frames and merge them into 32 image features. Experimental results indicate that this training method significantly outperforms the conventional uniform frame sampling technique with the same number of visual tokens.
\section{Experiments}

\begin{table*}[t]
  \centering
  \vspace{-0.1in}
  \scalebox{0.9}{ 
    \begin{tabular}{c|ccccccc}
      \toprule
      Method & VideoMME & Longvideobench & MLVU & MVBench & Egoschema &NextQA&PerceptionTest \\
      \midrule
      VideoChatGPT \cite{videochatgpt} & - & - & 31.3 & 32.7 & -&-&-\\
      TimeChat-7B \cite{timechat} & - & - & 30.9 & - & 33.0&-&- \\
      ShareGPT4Video-8B \cite{sharegpt4video} & 39.9/43.6 & 41.8 & 46.4 & 51.2 & -&-&-\\
      Video-LLaVA-7B \cite{videollava} & 39.9/41.6 & 37.6 & 47.3 & 43.5 & 38.4&- &-\\
      IXC-2.5-7B \cite{IXC-2.5} & 55.8/-&-&58.8&\textbf{69.1}&-&-&34.4 \\ 
      LLaVA-Next-Video-34B \cite{zhang2024llavanextvideo} & 52.0/54.9 & 50.5 & - & - & 49.3&70.2&51.6\\
      LongVA-7B \cite{long_va} & 52.6/- & - & 56.3 & - & -&68.3&-\\
      VideoLLaMA2-7B \cite{videollama2} & 47.9/50.3 & - & 32.7 & 54.6 & 51.7&-&51.4\\
      LLaVA-OV-7B \cite{llavaonevision} & 58.2/61.5 & 56.5 & 64.7 & 56.7 & 60.1&79.4&57.1\\
      Kangaroo-8B \cite{liu2024kangaroopowerfulvideolanguagemodel} & 56.0/57.6 & 54.8 & 61.0 & 61.1 & \textbf{62.7} &-&-\\
      \midrule
      FiLA-Base-7B & 61.5/63.4 & 55.9 & 67.6 & 52.9 & 49.8&81.1&67.1\\
      FiLA-Video-7B & \textbf{62.3/64.3} & \textbf{56.9} & \textbf{68.5} & 56.6 & 52.2&\textbf{81.8}&\textbf{67.9}\\
      \bottomrule
    \end{tabular}
  }
    \caption{Evaluation on several benchmarks with the state-of-art video understanding models. The FiLA-Base-7B, which uses uniform frame sampling in the video-only tuning stage, is trained on llava-video-178k dataset based on the stage-3 model same as FiLA-Video-7B, inferences with 64 frames. FiLA-Video-7B achieved the best performance on VideoMME, Longvideobench, MLVU, NextQA and PerceptionTest. Moreover, it achieves better results than FiLA-Base-7B with half number of visual tokens used by FiLA-Base-7B during reasoning.}
  \label{tab:study 1}
\end{table*}

\begin{table}[htp]
  \centering
  \vspace{-0.1in}
  \resizebox{0.4\textwidth}{!}{ 
    \begin{tabular}{c|cc}
      \toprule
      Method & VideoMME Long & MLVU \\
      \midrule
      Uniformly Sampling & 51.0 & 65.9\\
      BSM \ref{app:bsm} & 48.4 & \textbf{67.5} \\
      Ours (dissimilar) & 51.4 & 67.4 \\
      Ours (similar) & \textbf{51.6} & \textbf{67.5} \\
      \bottomrule
    \end{tabular}
  }
  \caption{Comparison of different implementation of scene selection on long video. All of them are only used during inferencing with the model trained on a 20\%-subset of llava-video-178K. BSM refers to the method using bipartite-soft-matching to do the selection. We also compare the result between selecting dissimilar frames as supplements and the opposite.}
  \label{tab:scene selection comp}
\end{table}

\begin{table}
  \centering
  \resizebox{0.5\textwidth}{!}{ 
    \begin{tabular}{c|cccc}
      \toprule
      Method & Short & Medium & Long & Overall \\
      \midrule
      Baseline(Uniformly Sampling) & 71.8 & 58.4 & 49.0 & 59.7\\
      \midrule
      Attention Pooling & 72.9 & 58.8 & 50.7 & 60.8 \\
      Temporal Averaging  & 73.4 & 59.3 & 50.2 & 61.0 \\
      BSM\cite{tome} & \textbf{73.6} & 59.4 & 49.4 & 60.8 \\
      Weighted Fusion & 73.1 & \textbf{59.4} & \textbf{51.0} & \textbf{61.2} \\
      \bottomrule
    \end{tabular}
  }
   \caption{The comparision of different merging methods on VideoMME without subtitles. All the models are trained on a 20\%-subset of llava-video-178K. }
  \label{tab:study 2}
\end{table}

\subsection{Details}

We use qwen2-7b-instruct \cite{qwen2} as the language model and siglip-so400m-patch14-384 \cite{siglip} as the vision encoder. An uninitialized 2-layer MLP with GELU is used as the projector. Each stage is trained for only one epoch.
In the first stage, we set the learning rate to 1e-3 and train only the projector.
In the second stage, we train the vision encoder, projector, and LLM with learning rates set to 2e-6, 2e-6, and 1e-5, respectively.
In the third stage, we continue to train the vision encoder, projector, and LLM with the same learning rates: 2e-6, 2e-6, and 1e-5.
In the fourth stage, we just focus on training the model’s merging ability at this stage. We uniformly sample 96 frames without scene-selection (We uniformly take every three frames as a scene) and merge them to 32 frames. We train the vision encoder, projector, LLM, and scene-merging module with learning rates set to 2e-6, 2e-6, 1e-5, and 2e-6, respectively.
Additionally, we use cosine scheduling.

\begin{table*}[htp]
    \centering

    \setlength{\tabcolsep}{6pt} 
    \renewcommand{\arraystretch}{1.2} 
    \scalebox{0.7}{%
    \begin{tabular}{@{}ll|c|c|c|c@{}}
    \toprule
    & & \textbf{Stage-1} & \textbf{Stage-2} & \textbf{Stage-3} & \textbf{Stage-4} \\
    \midrule 
    \multirow{2}{*}{\rotatebox[origin=c]{90}{\footnotesize \textit{Vision}}}
    & \textbf{Vision Encoder}  & \multicolumn{4}{c}{SigLip-so400m-patch14-384 \cite{siglip}} \\
    & \#Frames & - & - & - & FiLA-Base: 32, FiLA-Video: 96 merged to 32 \\
    \midrule 
    \multirow{4}{*}{\rotatebox[origin=c]{90}{\footnotesize \textit{Model}}}
    & \textbf{Language Model} & \multicolumn{4}{c}{Qwen2-7b-Instruct \cite{qwen2}} \\
    & \textbf{Projector} & \multicolumn{4}{c}{2-layer MLP with GELU} \\
    & Trainable Modules & Projector & Vision Encoder, Projector, LLM & Vision Encoder, Projector, LLM & Vision Encoder, Projector, LLM, Merging Module \\
    & \#Epochs & 1 & 1 & 1 & 1 \\
    \midrule 
    \multirow{6}{*}{\rotatebox[origin=c]{90}{\footnotesize \textit{Training}}}
    & \textbf{Learning Rate (Vision Encoder)} & - & 2 $\times 10^{-6}$ & 2 $\times 10^{-6}$ & 2 $\times 10^{-6}$ \\
    & \textbf{Learning Rate (Projector)} & 1 $\times 10^{-3}$ & 2 $\times 10^{-6}$ & 2 $\times 10^{-6}$ & 2 $\times 10^{-6}$ \\
    & \textbf{Learning Rate (LLM)} & - & 1 $\times 10^{-5}$ & 1 $\times 10^{-5}$ & 1 $\times 10^{-5}$ \\
    & \textbf{Learning Rate (Merging Module)} & - & - & - & 2 $\times 10^{-6}$ \\
    & \textbf{Batch Size} & 32 & 64 & 128 & 128 \\
    \bottomrule
    \end{tabular}
    }
    \vspace{1mm}
    \caption{Detailed configuration for each training stage of FiLA-Video-7B. The table outlines the model specifications and training hyperparameters across different stages of the learning process.}
    \vspace{3mm}
    \label{tab:training_strategy}
\end{table*}

\subsection{Evaluation}
\begin{figure*}[ht]  
  \centering
  \includegraphics[width=.9\textwidth]{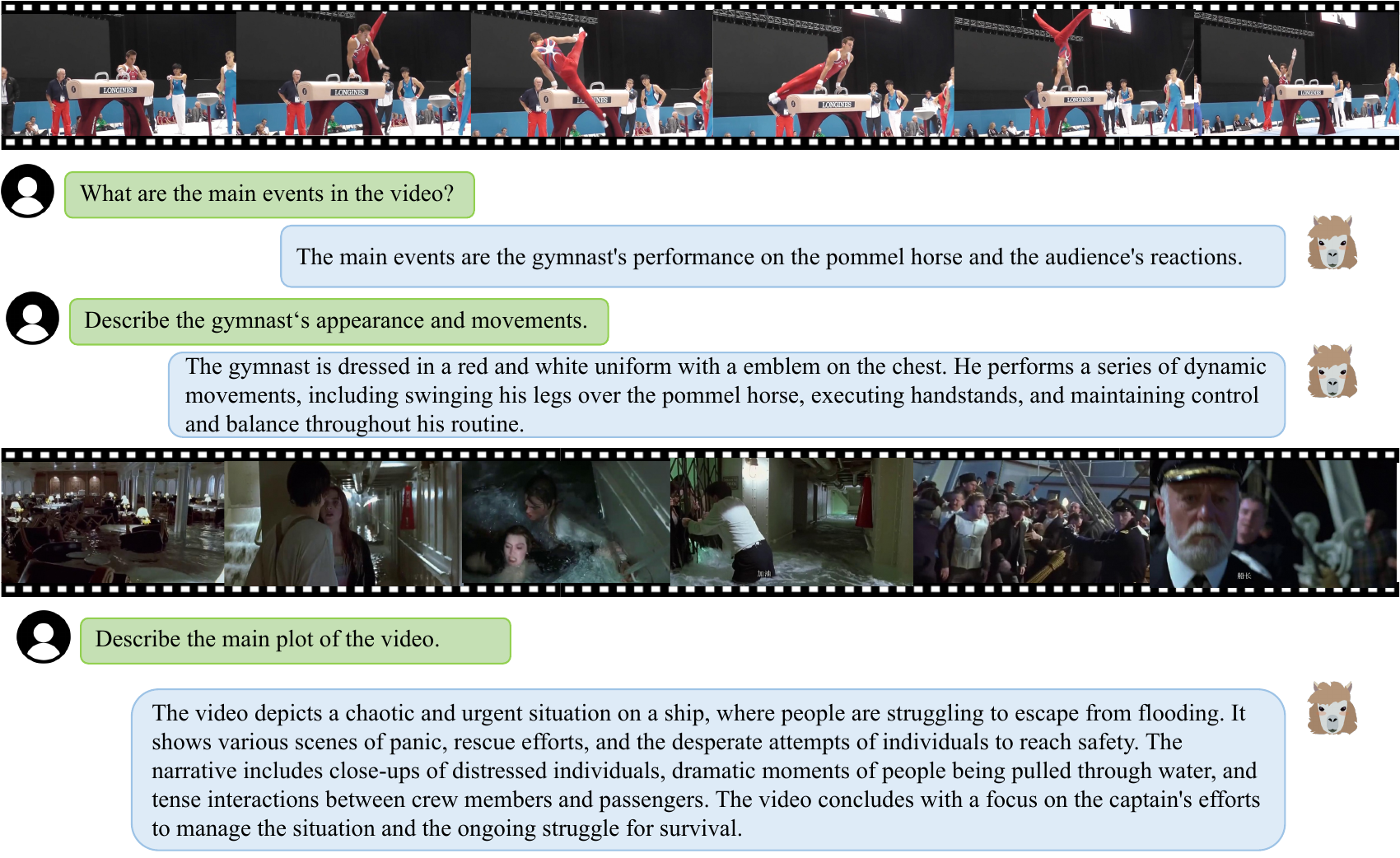} 
  \caption{Illustration of FiLA-video’s ability to conduct Q\&A based on video content. It can understand the main events from short videos like gymnastics and the main plot of long videos such as movie clips.}
  \label{fig:case} 
  \vspace{2mm}
\end{figure*}

Following prior work, we evaluated our FiLA-Video-7B on several popular benchmarks. Furthermore, we trained a baseline model, FiLA-Base-7B, for comparison, using the same training strategy and data as FiLA-Video. However, instead of employing the merging method, we simply sampled frames evenly for training. We also compared our model with some state-of-the-art multi-language large models (MLLMs) on VideoMME \cite{videomme}, MVBench \cite{mvbench}, LongVideoBench \cite{=longvideobenchbenchmarklongcontextinterleaved}, MLVU \cite{mlvu}, Egoschema \cite{egoschema}, NextQA \cite{nextqa}, and PerceptionTest \cite{perceptiontest}. The results can be seen in Table \ref{tab:study 1}. FiLA-Video-7B achieved the best performance on VideoMME, LongVideoBench, MLVU, NextQA, and PerceptionTest. It is notable that, compared to the baseline model FiLA-Base-7B, which uses 64 frames for reasoning, FiLA-Video-7B uses half the number of visual tokens during inference yet achieves better results. This demonstrates the effectiveness of our method in video feature fusion. As illustrated in Figure \ref{fig:case}, FiLA-Video-7B can comprehend different videos well and perform effectively on the task of Video-QA.

\subsection{Ablation study}
\textbf{Scene Selection} 
For the scene selection approach, we explored different implementation methods. One approach is scene selection based on BSM(bipartite soft matching) mentioned in \cite{tome}, with specific implementation details available in Appendix \ref{app:bsm}. The other approach is scene selection based on clustering methods, as described in the main text \ref{sceneselec}. We also explored the difference between selecting similar historical frames as supplements and selecting dissimilar frames as supplements after determining representative frames for the chosen scenes. Notably, we only introduced the scene selection module during inferencing. A comparison of the experimental results for these methods can be seen in Table \ref{tab:scene selection comp}. Finally, we choose the method based on clustering, selecting similar historical frames as supplements.

\begin{figure}[htbp]
  \centering
  \includegraphics[width=.5\textwidth]{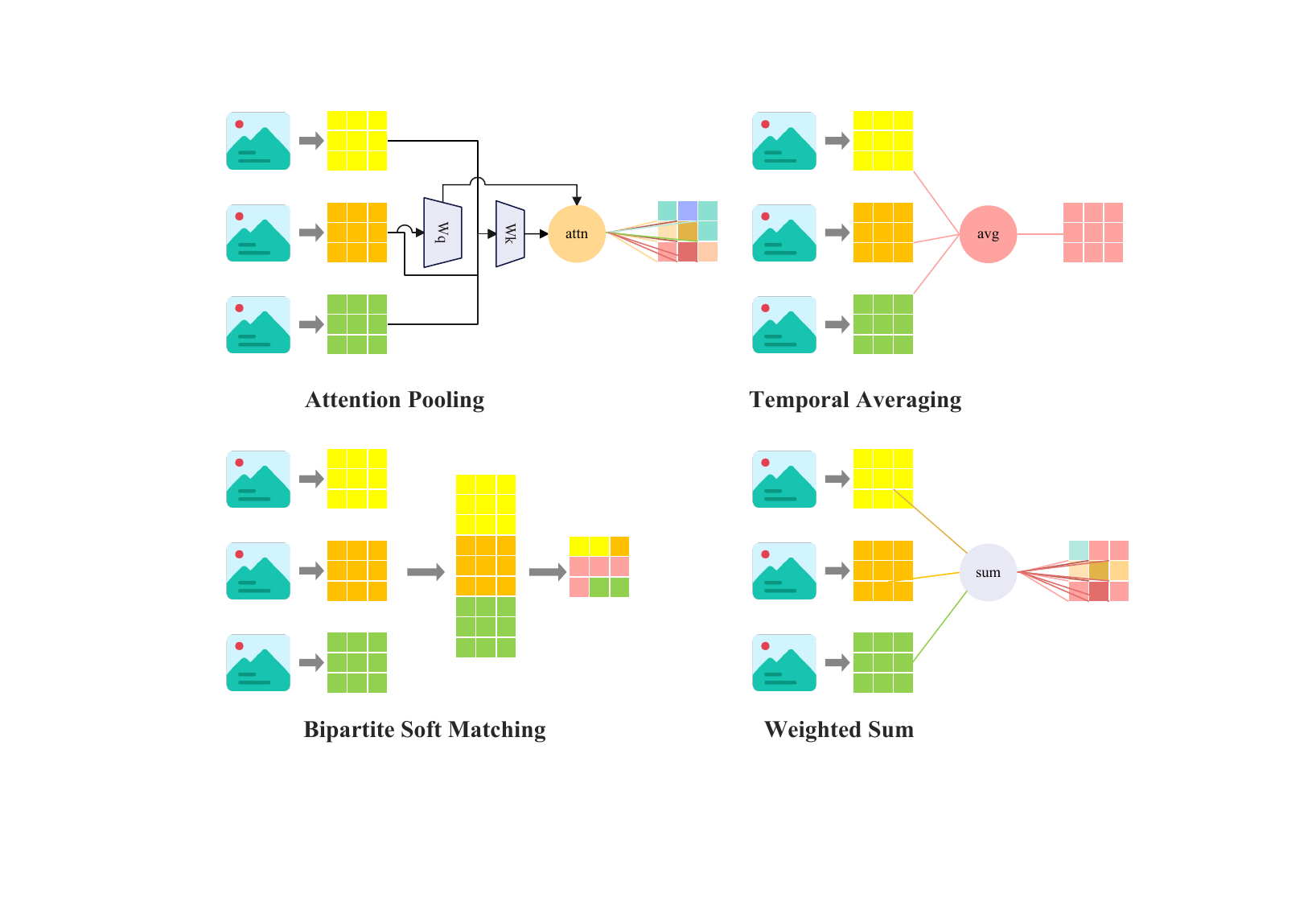} 
  \caption{Different merging methods.}
  \label{fig:methods} 
\end{figure}

\textbf{Scene Merging} 
As shown in Figure \ref{fig:methods}, we implemented four methods and compared them.
We implemented temporal averaging like \cite{tavg} and weighted fusion in Section \ref{method:merge}. More implementation details of other methods can be seen in Appendix \ref{app:methods} . We randomly sampled 20\% of the data from the llava-video-178k dataset in training stage 4 for the comparative experiments with the seed set to 42. The baseline model uses uniform sampling of 32 frames during training, while the other methods uniformly sample 96 frames and merge them into 32 frames. These methods include summing based on attention weights of the intermediate frames, uniform pooling in the time domain, merging based on bipartite soft matching, and using simple trainable weights. The result can be seen in Table \ref{tab:study 2}, which indicate that even without scene-selection, the merging methods can
outperform the uniform frame sampling technique with the same number of visual tokens sent to language model.

\begin{figure}[]
  \centering \includegraphics[width=.5\textwidth]{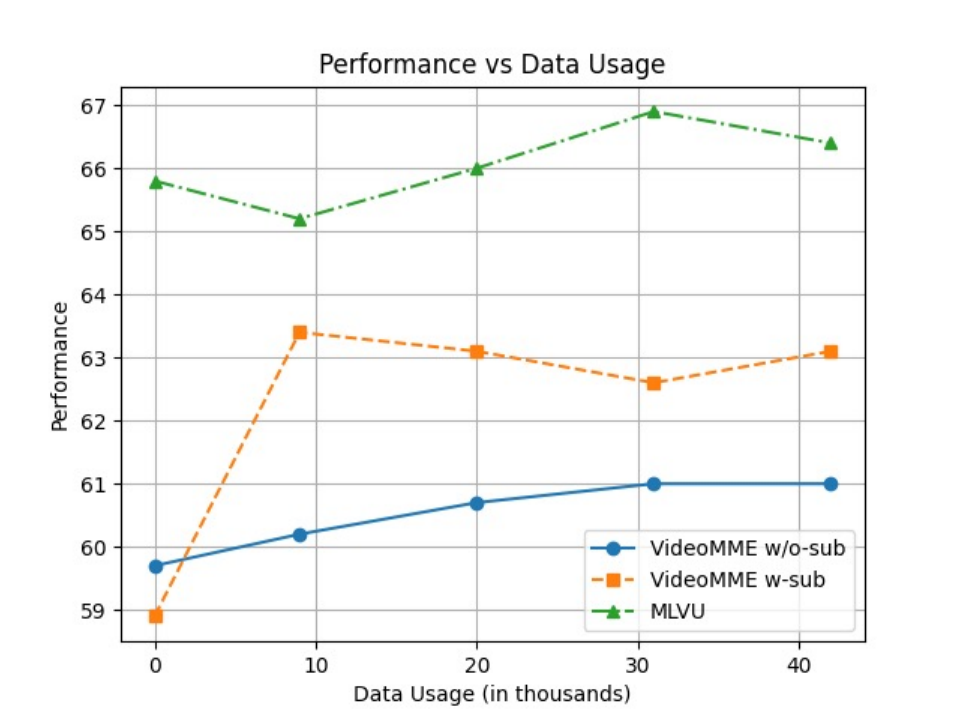} 
  \caption{The performance on several benchmarks changes as the amount of synthetic data changes. What can be seen from the result is that, except for the model trained with 9k synthetic data on MLVU, the models trained with synthetic data on these benchmarks are all higher than the models without synthetic data training.}
  \label{fig:line} 
\end{figure}

\begin{table}
  \centering
  \resizebox{0.5\textwidth}{!}{
  \begin{tabular}{c|ccc}
    \toprule
    Method & VideoMME w/o-sub&VideoMME w-sub  &MLVU \\
    \midrule
    lv178k-20\%-subset(334k) & 59.7&58.9 & 65.8\\
    \midrule
    lv178k-20\%-subset + 9k & 60.2(+0.5)&\textbf{63.4}(+4.5) &65.2(-0.6)\\
    lv178k-20\%-subset + 20k & 60.7(+1.0)&63.1(+4.2) &66.0(+0.2)\\
    lv178k-20\%-subset + 31k & \textbf{61.0}(+1.3)&62.6(+3.7)&\textbf{66.9}(+1.1) \\
    lv178k-20\%-subset + 42k & \textbf{61.0}(+1.3)&63.1(+4.2)&66.4(+0.6) \\
    \bottomrule
  \end{tabular}
  }
  \caption{The Comparision of different magnitudes of our synthetic data mixed with 20\%-subset of llava-video-178K. }
  \label{data}
\end{table}
\textbf{Synthetical Data} 
We randomly sampled 20\% of the data from the llava-video-178k dataset to train the baseline model for comparison, with the seed set to 42. For synthetic data, we mixed it into the sub-dataset for training with uniformly increasing magnitudes in our training stage 4. The experimental results can be seen in Table \ref{data}. As shown in Figure  \ref{fig:line}, we created a visual line chart to show the impact of synthetic data on several benchmarks, such as VideoMME and MLVU. The performance of the model trained with synthetic data on MLVU first decreases and then increases as the amount of data used increases. It remains higher than the baseline not trained with synthetic data on both VideoMME-with-subtitles and VideoMME-without-subtitles.

\section{Conclusions}
In this study, we propose FiLA-Video, a spatio-temporal feature compression method based on scene selection, aimed at enhancing the understanding of long videos. By selecting and merging key scenes in the video, we successfully reduce the number of video tokens to be processed while capturing more effective information. Our experimental results demonstrate that the proposed method performs well across multiple benchmarks, validating its effectiveness in long video understanding tasks. Additionally, we created a long-video visual captioning dataset to further strengthen the model's capability to understand long video content. This dataset has shown significant performance improvements on several benchmarks, proving its advantages in handling long-duration video datasets. Our research achieves a good balance between computational efficiency and context retention. Future work may focus on further optimizing scene selection and merging strategies to achieve more efficient feature representation and more accurate long video analysis.

\clearpage
 \small \bibliographystyle{ieeenat_fullname} \bibliography{main}

\begin{thebibliography}{67}
\providecommand{\natexlab}[1]{#1}
\providecommand{\url}[1]{\texttt{#1}}
\expandafter\ifx\csname urlstyle\endcsname\relax
  \providecommand{\doi}[1]{doi: #1}\else
  \providecommand{\doi}{doi: \begingroup \urlstyle{rm}\Url}\fi

\bibitem[Alayrac et~al.(2022)Alayrac, Donahue, Luc, Miech, Barr, Hasson, Lenc, Mensch, Millican, Reynolds, Ring, Rutherford, Cabi, Han, Gong, Samangooei, Monteiro, Menick, Borgeaud, Brock, Nematzadeh, Sharifzadeh, Binkowski, Barreira, Vinyals, Zisserman, and Simonyan]{mllm2alayrac2022flamingovisuallanguagemodel}
Jean-Baptiste Alayrac, Jeff Donahue, Pauline Luc, Antoine Miech, Iain Barr, Yana Hasson, Karel Lenc, Arthur Mensch, Katie Millican, Malcolm Reynolds, Roman Ring, Eliza Rutherford, Serkan Cabi, Tengda Han, Zhitao Gong, Sina Samangooei, Marianne Monteiro, Jacob Menick, Sebastian Borgeaud, Andrew Brock, Aida Nematzadeh, Sahand Sharifzadeh, Mikolaj Binkowski, Ricardo Barreira, Oriol Vinyals, Andrew Zisserman, and Karen Simonyan.
\newblock Flamingo: a visual language model for few-shot learning, 2022.

\bibitem[Bai et~al.(2023)Bai, Bai, Yang, Wang, Tan, Wang, Lin, Zhou, and Zhou]{mllm3bai2023qwenvlversatilevisionlanguagemodel}
Jinze Bai, Shuai Bai, Shusheng Yang, Shijie Wang, Sinan Tan, Peng Wang, Junyang Lin, Chang Zhou, and Jingren Zhou.
\newblock Qwen-vl: A versatile vision-language model for understanding, localization, text reading, and beyond, 2023.

\bibitem[Bolya et~al.(2023)Bolya, Fu, Dai, Zhang, Feichtenhofer, and Hoffman]{tome}
Daniel Bolya, Cheng-Yang Fu, Xiaoliang Dai, Peizhao Zhang, Christoph Feichtenhofer, and Judy Hoffman.
\newblock Token merging: Your vit but faster, 2023.

\bibitem[Chai et~al.(2024)Chai, Song, Du, Meng, Madhavan, Bar-Tal, Hwang, Xie, and Manning]{auroracap}
Wenhao Chai, Enxin Song, Yilun Du, Chenlin Meng, Vashisht Madhavan, Omer Bar-Tal, Jeng-Neng Hwang, Saining Xie, and Christopher~D. Manning.
\newblock Auroracap: Efficient, performant video detailed captioning and a new benchmark, 2024.

\bibitem[Chandrasegaran et~al.(2024)Chandrasegaran, Gupta, Hadzic, Kota, He, Eyzaguirre, Durante, Li, Wu, and Fei-Fei]{hourvideo}
Keshigeyan Chandrasegaran, Agrim Gupta, Lea~M. Hadzic, Taran Kota, Jimming He, Cristóbal Eyzaguirre, Zane Durante, Manling Li, Jiajun Wu, and Li Fei-Fei.
\newblock Hourvideo: 1-hour video-language understanding, 2024.

\bibitem[Chen et~al.(2024{\natexlab{a}})Chen, Wei, Li, Dong, Zhang, Zang, Chen, Duan, Lin, Tang, Yuan, Qiao, Lin, Zhao, and Wang]{sharegpt4video}
Lin Chen, Xilin Wei, Jinsong Li, Xiaoyi Dong, Pan Zhang, Yuhang Zang, Zehui Chen, Haodong Duan, Bin Lin, Zhenyu Tang, Li Yuan, Yu Qiao, Dahua Lin, Feng Zhao, and Jiaqi Wang.
\newblock Sharegpt4video: Improving video understanding and generation with better captions, 2024{\natexlab{a}}.

\bibitem[Chen et~al.(2024{\natexlab{b}})Chen, Zhao, Liu, Bai, Lin, Zhou, and Chang]{fastv}
Liang Chen, Haozhe Zhao, Tianyu Liu, Shuai Bai, Junyang Lin, Chang Zhou, and Baobao Chang.
\newblock An image is worth 1/2 tokens after layer 2: Plug-and-play inference acceleration for large vision-language models, 2024{\natexlab{b}}.

\bibitem[Chen et~al.(2024{\natexlab{c}})Chen, Siarohin, Menapace, Deyneka, wei Chao, Jeon, Fang, Lee, Ren, Yang, and Tulyakov]{panda70m}
Tsai-Shien Chen, Aliaksandr Siarohin, Willi Menapace, Ekaterina Deyneka, Hsiang wei Chao, Byung~Eun Jeon, Yuwei Fang, Hsin-Ying Lee, Jian Ren, Ming-Hsuan Yang, and Sergey Tulyakov.
\newblock Panda-70m: Captioning 70m videos with multiple cross-modality teachers, 2024{\natexlab{c}}.

\bibitem[Chen et~al.(2024{\natexlab{d}})Chen, Wu, Wang, Su, Chen, Xing, Zhong, Zhang, Zhu, Lu, Li, Luo, Lu, Qiao, and Dai]{mllm4chen2024internvlscalingvisionfoundation}
Zhe Chen, Jiannan Wu, Wenhai Wang, Weijie Su, Guo Chen, Sen Xing, Muyan Zhong, Qinglong Zhang, Xizhou Zhu, Lewei Lu, Bin Li, Ping Luo, Tong Lu, Yu Qiao, and Jifeng Dai.
\newblock Internvl: Scaling up vision foundation models and aligning for generic visual-linguistic tasks, 2024{\natexlab{d}}.

\bibitem[Cheng et~al.(2024)Cheng, Leng, Zhang, Xin, Li, Chen, Zhu, Zhang, Luo, Zhao, and Bing]{videollama2}
Zesen Cheng, Sicong Leng, Hang Zhang, Yifei Xin, Xin Li, Guanzheng Chen, Yongxin Zhu, Wenqi Zhang, Ziyang Luo, Deli Zhao, and Lidong Bing.
\newblock Videollama 2: Advancing spatial-temporal modeling and audio understanding in video-llms, 2024.

\bibitem[Endo et~al.(2024)Endo, Wang, and Yeung-Levy]{featherthrottle}
Mark Endo, Xiaohan Wang, and Serena Yeung-Levy.
\newblock Feather the throttle: Revisiting visual token pruning for vision-language model acceleration, 2024.

\bibitem[Fu et~al.(2024)Fu, Dai, Luo, Li, Ren, Zhang, Wang, Zhou, Shen, Zhang, Chen, Li, Lin, Zhao, Li, Xu, Zheng, Chen, Ji, and Sun]{videomme}
Chaoyou Fu, Yuhan Dai, Yongdong Luo, Lei Li, Shuhuai Ren, Renrui Zhang, Zihan Wang, Chenyu Zhou, Yunhang Shen, Mengdan Zhang, Peixian Chen, Yanwei Li, Shaohui Lin, Sirui Zhao, Ke Li, Tong Xu, Xiawu Zheng, Enhong Chen, Rongrong Ji, and Xing Sun.
\newblock Video-mme: The first-ever comprehensive evaluation benchmark of multi-modal llms in video analysis, 2024.

\bibitem[Grauman et~al.(2022)Grauman, Westbury, Byrne, Chavis, Furnari, Girdhar, Hamburger, Jiang, Liu, Liu, Martin, Nagarajan, Radosavovic, Ramakrishnan, Ryan, Sharma, Wray, Xu, Xu, Zhao, Bansal, Batra, Cartillier, Crane, Do, Doulaty, Erapalli, Feichtenhofer, Fragomeni, Fu, Gebreselasie, Gonzalez, Hillis, Huang, Huang, Jia, Khoo, Kolar, Kottur, Kumar, Landini, Li, Li, Li, Mangalam, Modhugu, Munro, Murrell, Nishiyasu, Price, Puentes, Ramazanova, Sari, Somasundaram, Southerland, Sugano, Tao, Vo, Wang, Wu, Yagi, Zhao, Zhu, Arbelaez, Crandall, Damen, Farinella, Fuegen, Ghanem, Ithapu, Jawahar, Joo, Kitani, Li, Newcombe, Oliva, Park, Rehg, Sato, Shi, Shou, Torralba, Torresani, Yan, and Malik]{ego4}
Kristen Grauman, Andrew Westbury, Eugene Byrne, Zachary Chavis, Antonino Furnari, Rohit Girdhar, Jackson Hamburger, Hao Jiang, Miao Liu, Xingyu Liu, Miguel Martin, Tushar Nagarajan, Ilija Radosavovic, Santhosh~Kumar Ramakrishnan, Fiona Ryan, Jayant Sharma, Michael Wray, Mengmeng Xu, Eric~Zhongcong Xu, Chen Zhao, Siddhant Bansal, Dhruv Batra, Vincent Cartillier, Sean Crane, Tien Do, Morrie Doulaty, Akshay Erapalli, Christoph Feichtenhofer, Adriano Fragomeni, Qichen Fu, Abrham Gebreselasie, Cristina Gonzalez, James Hillis, Xuhua Huang, Yifei Huang, Wenqi Jia, Weslie Khoo, Jachym Kolar, Satwik Kottur, Anurag Kumar, Federico Landini, Chao Li, Yanghao Li, Zhenqiang Li, Karttikeya Mangalam, Raghava Modhugu, Jonathan Munro, Tullie Murrell, Takumi Nishiyasu, Will Price, Paola~Ruiz Puentes, Merey Ramazanova, Leda Sari, Kiran Somasundaram, Audrey Southerland, Yusuke Sugano, Ruijie Tao, Minh Vo, Yuchen Wang, Xindi Wu, Takuma Yagi, Ziwei Zhao, Yunyi Zhu, Pablo Arbelaez, David Crandall, Dima Damen, Giovanni~Maria
  Farinella, Christian Fuegen, Bernard Ghanem, Vamsi~Krishna Ithapu, C.~V. Jawahar, Hanbyul Joo, Kris Kitani, Haizhou Li, Richard Newcombe, Aude Oliva, Hyun~Soo Park, James~M. Rehg, Yoichi Sato, Jianbo Shi, Mike~Zheng Shou, Antonio Torralba, Lorenzo Torresani, Mingfei Yan, and Jitendra Malik.
\newblock Ego4d: Around the world in 3,000 hours of egocentric video, 2022.

\bibitem[Heilbron et~al.(2015)Heilbron, Escorcia, Ghanem, and Niebles]{activitynet}
Fabian~Caba Heilbron, Victor Escorcia, Bernard Ghanem, and Juan~Carlos Niebles.
\newblock Activitynet: A large-scale video benchmark for human activity understanding.
\newblock In \emph{2015 IEEE Conference on Computer Vision and Pattern Recognition (CVPR)}, pages 961--970, 2015.

\bibitem[Hendricks et~al.(2017)Hendricks, Wang, Shechtman, Sivic, Darrell, and Russell]{didemo}
Lisa~Anne Hendricks, Oliver Wang, Eli Shechtman, Josef Sivic, Trevor Darrell, and Bryan Russell.
\newblock Localizing moments in video with natural language, 2017.

\bibitem[Huang et~al.(2024{\natexlab{a}})Huang, Zhang, Gao, Hu, and Qin]{fromimagetovideo}
Suyuan Huang, Haoxin Zhang, Yan Gao, Yao Hu, and Zengchang Qin.
\newblock From image to video, what do we need in multimodal llms?, 2024{\natexlab{a}}.

\bibitem[Huang et~al.(2024{\natexlab{b}})Huang, Zhou, and Han]{prunevid}
Xiaohu Huang, Hao Zhou, and Kai Han.
\newblock Prunevid: Visual token pruning for efficient video large language models, 2024{\natexlab{b}}.

\bibitem[Ikotun et~al.(2023)Ikotun, Ezugwu, Abualigah, Abuhaija, and Heming]{kmeans}
Abiodun~M. Ikotun, Absalom~E. Ezugwu, Laith Abualigah, Belal Abuhaija, and Jia Heming.
\newblock K-means clustering algorithms: A comprehensive review, variants analysis, and advances in the era of big data.
\newblock \emph{Inf. Sci.}, 622\penalty0 (C):\penalty0 178–210, 2023.

\bibitem[Jin et~al.(2024)Jin, Takanobu, Zhang, Cao, and Yuan]{Chat-UniVi}
Peng Jin, Ryuichi Takanobu, Wancai Zhang, Xiaochun Cao, and Li Yuan.
\newblock Chat-univi: Unified visual representation empowers large language models with image and video understanding, 2024.

\bibitem[Li et~al.(2024{\natexlab{a}})Li, Zhang, Guo, Zhang, Li, Zhang, Zhang, Zhang, Li, Liu, and Li]{llavaonevision}
Bo Li, Yuanhan Zhang, Dong Guo, Renrui Zhang, Feng Li, Hao Zhang, Kaichen Zhang, Peiyuan Zhang, Yanwei Li, Ziwei Liu, and Chunyuan Li.
\newblock Llava-onevision: Easy visual task transfer, 2024{\natexlab{a}}.

\bibitem[Li et~al.(2024{\natexlab{b}})Li, He, Wang, Li, Wang, Luo, Wang, Wang, and Qiao]{videochat}
KunChang Li, Yinan He, Yi Wang, Yizhuo Li, Wenhai Wang, Ping Luo, Yali Wang, Limin Wang, and Yu Qiao.
\newblock Videochat: Chat-centric video understanding, 2024{\natexlab{b}}.

\bibitem[Li et~al.(2024{\natexlab{c}})Li, Wang, He, Li, Wang, Liu, Wang, Xu, Chen, Luo, Wang, and Qiao]{mvbench}
Kunchang Li, Yali Wang, Yinan He, Yizhuo Li, Yi Wang, Yi Liu, Zun Wang, Jilan Xu, Guo Chen, Ping Luo, Limin Wang, and Yu Qiao.
\newblock Mvbench: A comprehensive multi-modal video understanding benchmark, 2024{\natexlab{c}}.

\bibitem[Li et~al.(2023)Li, Wang, and Jia]{llamavid}
Yanwei Li, Chengyao Wang, and Jiaya Jia.
\newblock Llama-vid: An image is worth 2 tokens in large language models, 2023.

\bibitem[Lin et~al.(2024)Lin, Ye, Zhu, Cui, Ning, Jin, and Yuan]{videollava}
Bin Lin, Yang Ye, Bin Zhu, Jiaxi Cui, Munan Ning, Peng Jin, and Li Yuan.
\newblock Video-llava: Learning united visual representation by alignment before projection, 2024.

\bibitem[Liu et~al.(2023{\natexlab{a}})Liu, Li, Wu, and Lee]{liu2023visualinstructiontuning}
Haotian Liu, Chunyuan Li, Qingyang Wu, and Yong~Jae Lee.
\newblock Visual instruction tuning, 2023{\natexlab{a}}.

\bibitem[Liu et~al.(2023{\natexlab{b}})Liu, Li, Wu, and Lee]{mllm1llava}
Haotian Liu, Chunyuan Li, Qingyang Wu, and Yong~Jae Lee.
\newblock Visual instruction tuning, 2023{\natexlab{b}}.

\bibitem[Liu et~al.(2024{\natexlab{a}})Liu, Li, Li, and Lee]{mllm5liu2024improvedbaselinesvisualinstruction}
Haotian Liu, Chunyuan Li, Yuheng Li, and Yong~Jae Lee.
\newblock Improved baselines with visual instruction tuning, 2024{\natexlab{a}}.

\bibitem[Liu et~al.(2024{\natexlab{b}})Liu, Wang, Ma, Wu, Ma, Wei, Jiao, Wu, and Hu]{liu2024kangaroopowerfulvideolanguagemodel}
Jiajun Liu, Yibing Wang, Hanghang Ma, Xiaoping Wu, Xiaoqi Ma, Xiaoming Wei, Jianbin Jiao, Enhua Wu, and Jie Hu.
\newblock Kangaroo: A powerful video-language model supporting long-context video input, 2024{\natexlab{b}}.

\bibitem[Liu et~al.(2024{\natexlab{c}})Liu, Tang, Liu, Ge, Shan, Li, and Yang]{ppllava}
Ruyang Liu, Haoran Tang, Haibo Liu, Yixiao Ge, Ying Shan, Chen Li, and Jiankun Yang.
\newblock Ppllava: Varied video sequence understanding with prompt guidance, 2024{\natexlab{c}}.

\bibitem[Liu et~al.(2024{\natexlab{d}})Liu, Li, Liu, Wang, Ren, Li, Chen, Sun, and Hou]{tempcompass}
Yuanxin Liu, Shicheng Li, Yi Liu, Yuxiang Wang, Shuhuai Ren, Lei Li, Sishuo Chen, Xu Sun, and Lu Hou.
\newblock Tempcompass: Do video llms really understand videos?, 2024{\natexlab{d}}.

\bibitem[Liu et~al.(2024{\natexlab{e}})Liu, Zhu, Shi, Zhang, Lou, Yang, Xi, Cao, Gu, Li, Li, Fang, Chen, Hsieh, Huang, Cheng, Nath, Hu, Liu, Krishna, Xu, Wang, Molchanov, Kautz, Yin, Han, and Lu]{nvila}
Zhijian Liu, Ligeng Zhu, Baifeng Shi, Zhuoyang Zhang, Yuming Lou, Shang Yang, Haocheng Xi, Shiyi Cao, Yuxian Gu, Dacheng Li, Xiuyu Li, Yunhao Fang, Yukang Chen, Cheng-Yu Hsieh, De-An Huang, An-Chieh Cheng, Vishwesh Nath, Jinyi Hu, Sifei Liu, Ranjay Krishna, Daguang Xu, Xiaolong Wang, Pavlo Molchanov, Jan Kautz, Hongxu Yin, Song Han, and Yao Lu.
\newblock Nvila: Efficient frontier visual language models, 2024{\natexlab{e}}.

\bibitem[Lu et~al.(2024)Lu, Yin, He, Wang, Liu, Wang, and Hu]{bvllm}
Zhuqiang Lu, Zhenfei Yin, Mengwei He, Zhihui Wang, Zicheng Liu, Zhiyong Wang, and Kun Hu.
\newblock B-vllm: A vision large language model with balanced spatio-temporal tokens, 2024.

\bibitem[Luo et~al.(2023)Luo, Zhao, Yang, Dong, Li, Lu, Wang, Hu, Qiu, and Wei]{valley}
Ruipu Luo, Ziwang Zhao, Min Yang, Junwei Dong, Da Li, Pengcheng Lu, Tao Wang, Linmei Hu, Minghui Qiu, and Zhongyu Wei.
\newblock Valley: Video assistant with large language model enhanced ability, 2023.

\bibitem[Maaz et~al.(2024)Maaz, Rasheed, Khan, and Khan]{videochatgpt}
Muhammad Maaz, Hanoona Rasheed, Salman Khan, and Fahad~Shahbaz Khan.
\newblock Video-chatgpt: Towards detailed video understanding via large vision and language models, 2024.

\bibitem[Mangalam et~al.(2023)Mangalam, Akshulakov, and Malik]{egoschema}
Karttikeya Mangalam, Raiymbek Akshulakov, and Jitendra Malik.
\newblock Egoschema: A diagnostic benchmark for very long-form video language understanding, 2023.

\bibitem[Miech et~al.(2019)Miech, Zhukov, Alayrac, Tapaswi, Laptev, and Sivic]{howto100m}
Antoine Miech, Dimitri Zhukov, Jean-Baptiste Alayrac, Makarand Tapaswi, Ivan Laptev, and Josef Sivic.
\newblock Howto100m: Learning a text-video embedding by watching hundred million narrated video clips, 2019.

\bibitem[Ning et~al.(2023)Ning, Zhu, Xie, Lin, Cui, Yuan, Chen, and Yuan]{videobench}
Munan Ning, Bin Zhu, Yujia Xie, Bin Lin, Jiaxi Cui, Lu Yuan, Dongdong Chen, and Li Yuan.
\newblock Video-bench: A comprehensive benchmark and toolkit for evaluating video-based large language models, 2023.

\bibitem[Pătrăucean et~al.(2023)Pătrăucean, Smaira, Gupta, Continente, Markeeva, Banarse, Koppula, Heyward, Malinowski, Yang, Doersch, Matejovicova, Sulsky, Miech, Frechette, Klimczak, Koster, Zhang, Winkler, Aytar, Osindero, Damen, Zisserman, and Carreira]{perceptiontest}
Viorica Pătrăucean, Lucas Smaira, Ankush Gupta, Adrià~Recasens Continente, Larisa Markeeva, Dylan Banarse, Skanda Koppula, Joseph Heyward, Mateusz Malinowski, Yi Yang, Carl Doersch, Tatiana Matejovicova, Yury Sulsky, Antoine Miech, Alex Frechette, Hanna Klimczak, Raphael Koster, Junlin Zhang, Stephanie Winkler, Yusuf Aytar, Simon Osindero, Dima Damen, Andrew Zisserman, and João Carreira.
\newblock Perception test: A diagnostic benchmark for multimodal video models, 2023.

\bibitem[Radford et~al.(2021)Radford, Kim, Hallacy, Ramesh, Goh, Agarwal, Sastry, Askell, Mishkin, Clark, Krueger, and Sutskever]{clip}
Alec Radford, Jong~Wook Kim, Chris Hallacy, Aditya Ramesh, Gabriel Goh, Sandhini Agarwal, Girish Sastry, Amanda Askell, Pamela Mishkin, Jack Clark, Gretchen Krueger, and Ilya Sutskever.
\newblock Learning transferable visual models from natural language supervision, 2021.

\bibitem[Rawal et~al.(2024)Rawal, Saifullah, Farré, Basri, Jacobs, Somepalli, and Goldstein]{cinepile}
Ruchit Rawal, Khalid Saifullah, Miquel Farré, Ronen Basri, David Jacobs, Gowthami Somepalli, and Tom Goldstein.
\newblock Cinepile: A long video question answering dataset and benchmark, 2024.

\bibitem[Ren et~al.(2024)Ren, Yao, Li, Sun, and Hou]{timechat}
Shuhuai Ren, Linli Yao, Shicheng Li, Xu Sun, and Lu Hou.
\newblock Timechat: A time-sensitive multimodal large language model for long video understanding, 2024.

\bibitem[Rohrbach et~al.(2015)Rohrbach, Rohrbach, Tandon, and Schiele]{lsmdc}
Anna Rohrbach, Marcus Rohrbach, Niket Tandon, and Bernt Schiele.
\newblock A dataset for movie description, 2015.

\bibitem[Tao et~al.(2024)Tao, Qin, You, Sui, and Wang]{dycoke}
Keda Tao, Can Qin, Haoxuan You, Yang Sui, and Huan Wang.
\newblock Dycoke: Dynamic compression of tokens for fast video large language models, 2024.

\bibitem[Wang et~al.(2016)Wang, Xiong, Wang, Qiao, Lin, Tang, and Gool]{tavg}
Limin Wang, Yuanjun Xiong, Zhe Wang, Yu Qiao, Dahua Lin, Xiaoou Tang, and Luc~Van Gool.
\newblock Temporal segment networks: Towards good practices for deep action recognition, 2016.

\bibitem[Wang et~al.(2024{\natexlab{a}})Wang, Bai, Tan, Wang, Fan, Bai, Chen, Liu, Wang, Ge, Fan, Dang, Du, Ren, Men, Liu, Zhou, Zhou, and Lin]{qwen2-vl}
Peng Wang, Shuai Bai, Sinan Tan, Shijie Wang, Zhihao Fan, Jinze Bai, Keqin Chen, Xuejing Liu, Jialin Wang, Wenbin Ge, Yang Fan, Kai Dang, Mengfei Du, Xuancheng Ren, Rui Men, Dayiheng Liu, Chang Zhou, Jingren Zhou, and Junyang Lin.
\newblock Qwen2-vl: Enhancing vision-language model's perception of the world at any resolution, 2024{\natexlab{a}}.

\bibitem[Wang et~al.(2020)Wang, Wu, Chen, Li, Wang, and Wang]{vatex}
Xin Wang, Jiawei Wu, Junkun Chen, Lei Li, Yuan-Fang Wang, and William~Yang Wang.
\newblock Vatex: A large-scale, high-quality multilingual dataset for video-and-language research, 2020.

\bibitem[Wang et~al.(2024{\natexlab{b}})Wang, He, Li, Li, Yu, Ma, Li, Chen, Chen, Wang, He, Luo, Liu, Wang, Wang, and Qiao]{internvid}
Yi Wang, Yinan He, Yizhuo Li, Kunchang Li, Jiashuo Yu, Xin Ma, Xinhao Li, Guo Chen, Xinyuan Chen, Yaohui Wang, Conghui He, Ping Luo, Ziwei Liu, Yali Wang, Limin Wang, and Yu Qiao.
\newblock Internvid: A large-scale video-text dataset for multimodal understanding and generation, 2024{\natexlab{b}}.

\bibitem[Weng et~al.(2024)Weng, Han, He, Chang, and Zhuang]{longvlm}
Yuetian Weng, Mingfei Han, Haoyu He, Xiaojun Chang, and Bohan Zhuang.
\newblock Longvlm: Efficient long video understanding via large language models, 2024.

\bibitem[Wu et~al.(2024)Wu, Li, Chen, and Li]{=longvideobenchbenchmarklongcontextinterleaved}
Haoning Wu, Dongxu Li, Bei Chen, and Junnan Li.
\newblock Longvideobench: A benchmark for long-context interleaved video-language understanding, 2024.

\bibitem[Xiao et~al.(2021)Xiao, Shang, Yao, and Chua]{nextqa}
Junbin Xiao, Xindi Shang, Angela Yao, and Tat-Seng Chua.
\newblock Next-qa:next phase of question-answering to explaining temporal actions, 2021.

\bibitem[Xu et~al.(2016)Xu, Mei, Yao, and Rui]{msrvtt}
Jun Xu, Tao Mei, Ting Yao, and Yong Rui.
\newblock Msr-vtt: A large video description dataset for bridging video and language.
\newblock In \emph{2016 IEEE Conference on Computer Vision and Pattern Recognition (CVPR)}, pages 5288--5296, 2016.

\bibitem[Xue et~al.(2022)Xue, Hang, Zeng, Sun, Liu, Yang, Fu, and Guo]{hdvila100m}
Hongwei Xue, Tiankai Hang, Yanhong Zeng, Yuchong Sun, Bei Liu, Huan Yang, Jianlong Fu, and Baining Guo.
\newblock Advancing high-resolution video-language representation with large-scale video transcriptions, 2022.

\bibitem[Yang et~al.(2024)Yang, Yang, Hui, Zheng, Yu, Zhou, Li, Li, Liu, Huang, Dong, Wei, Lin, Tang, Wang, Yang, Tu, Zhang, Ma, Yang, Xu, Zhou, Bai, He, Lin, Dang, Lu, Chen, Yang, Li, Xue, Ni, Zhang, Wang, Peng, Men, Gao, Lin, Wang, Bai, Tan, Zhu, Li, Liu, Ge, Deng, Zhou, Ren, Zhang, Wei, Ren, Liu, Fan, Yao, Zhang, Wan, Chu, Liu, Cui, Zhang, Guo, and Fan]{qwen2}
An Yang, Baosong Yang, Binyuan Hui, Bo Zheng, Bowen Yu, Chang Zhou, Chengpeng Li, Chengyuan Li, Dayiheng Liu, Fei Huang, Guanting Dong, Haoran Wei, Huan Lin, Jialong Tang, Jialin Wang, Jian Yang, Jianhong Tu, Jianwei Zhang, Jianxin Ma, Jianxin Yang, Jin Xu, Jingren Zhou, Jinze Bai, Jinzheng He, Junyang Lin, Kai Dang, Keming Lu, Keqin Chen, Kexin Yang, Mei Li, Mingfeng Xue, Na Ni, Pei Zhang, Peng Wang, Ru Peng, Rui Men, Ruize Gao, Runji Lin, Shijie Wang, Shuai Bai, Sinan Tan, Tianhang Zhu, Tianhao Li, Tianyu Liu, Wenbin Ge, Xiaodong Deng, Xiaohuan Zhou, Xingzhang Ren, Xinyu Zhang, Xipin Wei, Xuancheng Ren, Xuejing Liu, Yang Fan, Yang Yao, Yichang Zhang, Yu Wan, Yunfei Chu, Yuqiong Liu, Zeyu Cui, Zhenru Zhang, Zhifang Guo, and Zhihao Fan.
\newblock Qwen2 technical report, 2024.

\bibitem[Yu et~al.(2023)Yu, Cho, Yadav, and Bansal]{self-chained}
Shoubin Yu, Jaemin Cho, Prateek Yadav, and Mohit Bansal.
\newblock Self-chained image-language model for video localization and question answering, 2023.

\bibitem[Zhai et~al.(2023)Zhai, Mustafa, Kolesnikov, and Beyer]{siglip}
Xiaohua Zhai, Basil Mustafa, Alexander Kolesnikov, and Lucas Beyer.
\newblock Sigmoid loss for language image pre-training, 2023.

\bibitem[Zhang et~al.(2025)Zhang, Li, Cheng, Hu, Yuan, Chen, Leng, Jiang, Zhang, Li, Jin, Zhang, Wang, Bing, and Zhao]{zhang2025videollama3frontiermultimodal}
Boqiang Zhang, Kehan Li, Zesen Cheng, Zhiqiang Hu, Yuqian Yuan, Guanzheng Chen, Sicong Leng, Yuming Jiang, Hang Zhang, Xin Li, Peng Jin, Wenqi Zhang, Fan Wang, Lidong Bing, and Deli Zhao.
\newblock Videollama 3: Frontier multimodal foundation models for image and video understanding, 2025.

\bibitem[Zhang et~al.(2023)Zhang, Li, and Bing]{videollama}
Hang Zhang, Xin Li, and Lidong Bing.
\newblock Video-llama: An instruction-tuned audio-visual language model for video understanding, 2023.

\bibitem[Zhang et~al.(2024{\natexlab{a}})Zhang, Dong, Zang, Cao, Qian, Chen, Guo, Duan, Wang, Ouyang, Zhang, Zhang, Li, Gao, Sun, Zhang, Li, Li, Wang, Yan, He, Zhang, Chen, Dai, Qiao, Lin, and Wang]{IXC-2.5}
Pan Zhang, Xiaoyi Dong, Yuhang Zang, Yuhang Cao, Rui Qian, Lin Chen, Qipeng Guo, Haodong Duan, Bin Wang, Linke Ouyang, Songyang Zhang, Wenwei Zhang, Yining Li, Yang Gao, Peng Sun, Xinyue Zhang, Wei Li, Jingwen Li, Wenhai Wang, Hang Yan, Conghui He, Xingcheng Zhang, Kai Chen, Jifeng Dai, Yu Qiao, Dahua Lin, and Jiaqi Wang.
\newblock Internlm-xcomposer-2.5: A versatile large vision language model supporting long-contextual input and output, 2024{\natexlab{a}}.

\bibitem[Zhang et~al.(2024{\natexlab{b}})Zhang, Zhang, Li, Zeng, Yang, Zhang, Wang, Tan, Li, and Liu]{long_va}
Peiyuan Zhang, Kaichen Zhang, Bo Li, Guangtao Zeng, Jingkang Yang, Yuanhan Zhang, Ziyue Wang, Haoran Tan, Chunyuan Li, and Ziwei Liu.
\newblock Long context transfer from language to vision, 2024{\natexlab{b}}.

\bibitem[Zhang et~al.(2024{\natexlab{c}})Zhang, Li, Liu, Lee, Gui, Fu, Feng, Liu, and Li]{zhang2024llavanextvideo}
Yuanhan Zhang, Bo Li, haotian Liu, Yong~jae Lee, Liangke Gui, Di Fu, Jiashi Feng, Ziwei Liu, and Chunyuan Li.
\newblock Llava-next: A strong zero-shot video understanding model, 2024{\natexlab{c}}.

\bibitem[Zhang et~al.(2024{\natexlab{d}})Zhang, Wu, Li, Li, Ma, Liu, and Li]{llava-video}
Yuanhan Zhang, Jinming Wu, Wei Li, Bo Li, Zejun Ma, Ziwei Liu, and Chunyuan Li.
\newblock Video instruction tuning with synthetic data, 2024{\natexlab{d}}.

\bibitem[Zhang et~al.(2024{\natexlab{e}})Zhang, Zhao, Chen, Ding, Yang, and Sun]{dyto}
Yiming Zhang, Zhuokai Zhao, Zhaorun Chen, Zenghui Ding, Xianjun Yang, and Yining Sun.
\newblock Beyond training: Dynamic token merging for zero-shot video understanding, 2024{\natexlab{e}}.

\bibitem[Zheng et~al.(2024)Zheng, Wang, Xie, Liu, Sun, Xin, Shen, Li, and Li]{lyra}
Chuanyang Zheng, Haiming Wang, Enze Xie, Zhengying Liu, Jiankai Sun, Huajian Xin, Jianhao Shen, Zhenguo Li, and Yu Li.
\newblock Lyra: Orchestrating dual correction in automated theorem proving, 2024.

\bibitem[Zhou et~al.(2025)Zhou, Shu, Zhao, Wu, Liang, Xiao, Qin, Yang, Xiong, Zhang, Huang, and Liu]{mlvu}
Junjie Zhou, Yan Shu, Bo Zhao, Boya Wu, Zhengyang Liang, Shitao Xiao, Minghao Qin, Xi Yang, Yongping Xiong, Bo Zhang, Tiejun Huang, and Zheng Liu.
\newblock Mlvu: Benchmarking multi-task long video understanding, 2025.

\bibitem[Zhou et~al.(2017)Zhou, Xu, and Corso]{youcook2}
Luowei Zhou, Chenliang Xu, and Jason~J. Corso.
\newblock Towards automatic learning of procedures from web instructional videos, 2017.

\bibitem[Zhu et~al.(2023{\natexlab{a}})Zhu, Chen, Shen, Li, and Elhoseiny]{mllm6zhu2023minigpt4enhancingvisionlanguageunderstanding}
Deyao Zhu, Jun Chen, Xiaoqian Shen, Xiang Li, and Mohamed Elhoseiny.
\newblock Minigpt-4: Enhancing vision-language understanding with advanced large language models, 2023{\natexlab{a}}.

\bibitem[Zhu et~al.(2023{\natexlab{b}})Zhu, Chen, Shen, Li, and Elhoseiny]{zhu2023minigpt4enhancingvisionlanguageunderstanding}
Deyao Zhu, Jun Chen, Xiaoqian Shen, Xiang Li, and Mohamed Elhoseiny.
\newblock Minigpt-4: Enhancing vision-language understanding with advanced large language models, 2023{\natexlab{b}}.

\end{thebibliography}

\clearpage
\setcounter{page}{1}
\maketitlesupplementary

\appendix
\setcounter{equation}{0}
\setcounter{table}{0}
\setcounter{figure}{0}

\section{Dataset}
The comparison of our dataset and others video-language datasets is shown in Figure \ref{tab:dataset}.

\section{bipartite soft matching for scene-selection}
\label{app:bsm}
In this comparative method, we follow the idea of bipartite soft matching from ToMe \cite{tome} for scene selection. First, for the input \(N\) frames, we evenly divide them into \(k\) segments. After calculating the representative features for each frame using the method outlined in Equation \ref{eq:mean}, we divide each segment into two roughly equal sets \(A\) and \(B\). We then draw \(r\) edges from each feature in \(A\) to \(r\) features in \(B\). Next, we keep one group of the most similar edges. Finally, we organize these retained frames in chronological order. By adjusting \(k\) and \(r\), we ensure that the number of frames is the same for both methods.
\section{merging methods}
\label{app:methods}
\subsection{attention pooling}
For \(M\) feature maps within a scene, we use the feature map of the middle frame as the query to compute cross-attention scores with the feature maps of other frames, using these scores as weights for each token to perform a weighted sum. For the projection matrices, we use Xavier initialization. 

Given a feature map \(\mathbf{F}_m\), the query \(\mathbf{Q}\) and key \(\mathbf{K}_m\) are obtained by projecting \(\mathbf{F}_m\) with the matrices \(\mathbf{W}_Q\) and \(\mathbf{W}_K\) respectively:

\begin{equation}
\mathbf{Q} = \mathbf{F}_{\text{query}} \mathbf{W}_Q
\end{equation}

\begin{equation}
\mathbf{K}_m = \mathbf{F}_m \mathbf{W}_K
\end{equation}

where \(\mathbf{W}_Q\) and \(\mathbf{W}_K\) are the projection matrices for the query and the keys, respectively.

The attention weight matrix \(\mathbf{A}_m\) is computed as:

\begin{equation}
\mathbf{A}_m = \text{softmax}\left(\frac{\mathbf{Q} \mathbf{K}_m^\top}{\sqrt{d_k}}\right)
\end{equation}

where \(d_k\) is the dimension of the key feature map.

The combined feature map \(\mathbf{F}_c\) is calculated by performing a weighted sum of each feature map \(\mathbf{F}_m\) multiplied by its respective attention weights \(\mathbf{A}_m\):

\begin{equation}
\mathbf{F}_c = \sum_{m=1}^{M} \mathbf{A}_m \cdot \mathbf{F}_m
\end{equation}

where the element-wise multiplication \(\cdot\) applies the attention weights \(\mathbf{A}_m\) to the feature map \(\mathbf{F}_m\).

\subsection{bipartite-soft-match merge}
Similar implementation have been used in \cite{longvlm}. In order to facilitate comparison, we have set corresponding limits on the number of merged target tokens. The pseudo code is showed in algorithm \ref{alg:bsm}.

\begin{algorithm}[]
\caption{Bipartite Soft Matching and Merging}
\label{alg:bsm}
\SetAlgoLined
\KwIn{Feature matrix $\mathbf{X}$, target number of tokens $T$}
\KwOut{Merged feature matrix $\mathbf{X}_{\text{merged}}$}

Initialize metric tensor $\mathbf{M}$ based on $\mathbf{X}$ \\
Initialize size tensor $\mathbf{S}$ as ones \\
Normalize $\mathbf{M}$: $\mathbf{M} \leftarrow \mathbf{M} / \|\mathbf{M}\|$ \\
Compute total tokens to remove $R \leftarrow \text{size}(\mathbf{X}, 1) - T$ \\

\While{$R > 0$} {
    Split metric tensor into $\mathbf{A}$ and $\mathbf{B}$: \\
    \Indp $\mathbf{A}, \mathbf{B} \leftarrow \mathbf{M}_{\cdots, ::2, \cdot}, \mathbf{M}_{\cdots, 1::2, \cdot}$ \\
    \Indm Compute attention scores $\mathbf{S} \leftarrow \mathbf{A} \cdot \mathbf{B}^\top$ \\
    Find maximum indices: $(\mathbf{v}_{\text{max}}, \mathbf{v}_{\text{idx}}) \leftarrow \max(\mathbf{S}, \text{dim} = -1)$ \\
    Sort edges by score: $\mathbf{e}_{\text{idx}} \leftarrow \text{argsort}(\mathbf{v}_{\text{max}}, \text{descending})$ \\
    Select merged and unmerged indices: \\
    \Indp $\mathbf{e}_{\text{unmerged}}, \mathbf{e}_{\text{merged}} \leftarrow \mathbf{e}_{{\text{idx}}_{\cdots, r:, \cdots}}, \mathbf{e}_{{\text{idx}}_{\cdots, :r, \cdots}}$ \\
    $\mathbf{t}_{\text{dst}} \leftarrow \mathbf{v}_{{\text{idx}}_{\cdots, \mathbf{e}_{\text{merged}}}}$ \\
    \Indm Gather and merge tokens:\\
    \Indp $\mathbf{S} \leftarrow \mathbf{X}_{\cdots, ::2, \cdot}$ \\
    $\mathbf{D} \leftarrow \mathbf{X}_{\cdots, 1::2, \cdot}$ \\
    $\mathbf{U} \leftarrow \mathbf{S}_{{\mathbf{e}_{\text{unmerged}}}}$ \\
    $\mathbf{S}_{\text{merge}} \leftarrow \mathbf{S}_{{\mathbf{e}_{\text{merged}}}}$ \\
    $\mathbf{D} \leftarrow \mathbf{D} + \mathbf{D}_{{\mathbf{t}_{\text{dst}}}}$ \\
    $\mathbf{X} \leftarrow \text{concat}(\mathbf{U}, \mathbf{D}, \text{dim} = 1)$ \\
    \Indm Update metric tensor $\mathbf{M}$ \\
    Update remaining tokens $R$ \\
}
Reshape $\mathbf{X}$ to target dimensions $\mathbf{X}_{\text{merged}} \leftarrow \mathbf{X}[1, T, \cdot]$ \\
\KwRet $\mathbf{X}_{\text{merged}}$

\end{algorithm}

\section{Case Study}
\begin{figure*}[t]
  \centering
  \begin{subfigure}[t]{\textwidth}
    \centering
    \includegraphics[width=.9\textwidth]{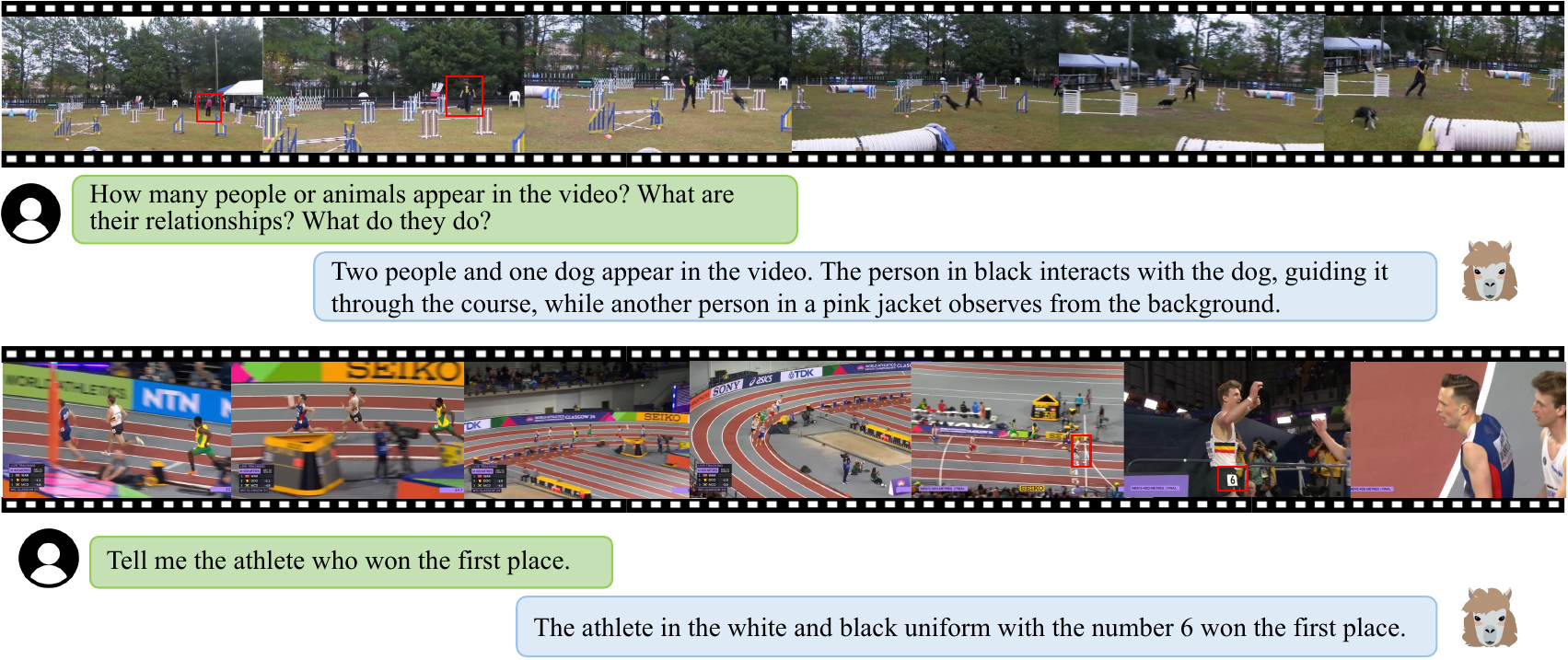}
    \label{fig:image1}
  \end{subfigure}

  \vskip\floatsep 

  \begin{subfigure}[t]{\textwidth}
    \centering
    \includegraphics[width=.9\textwidth]{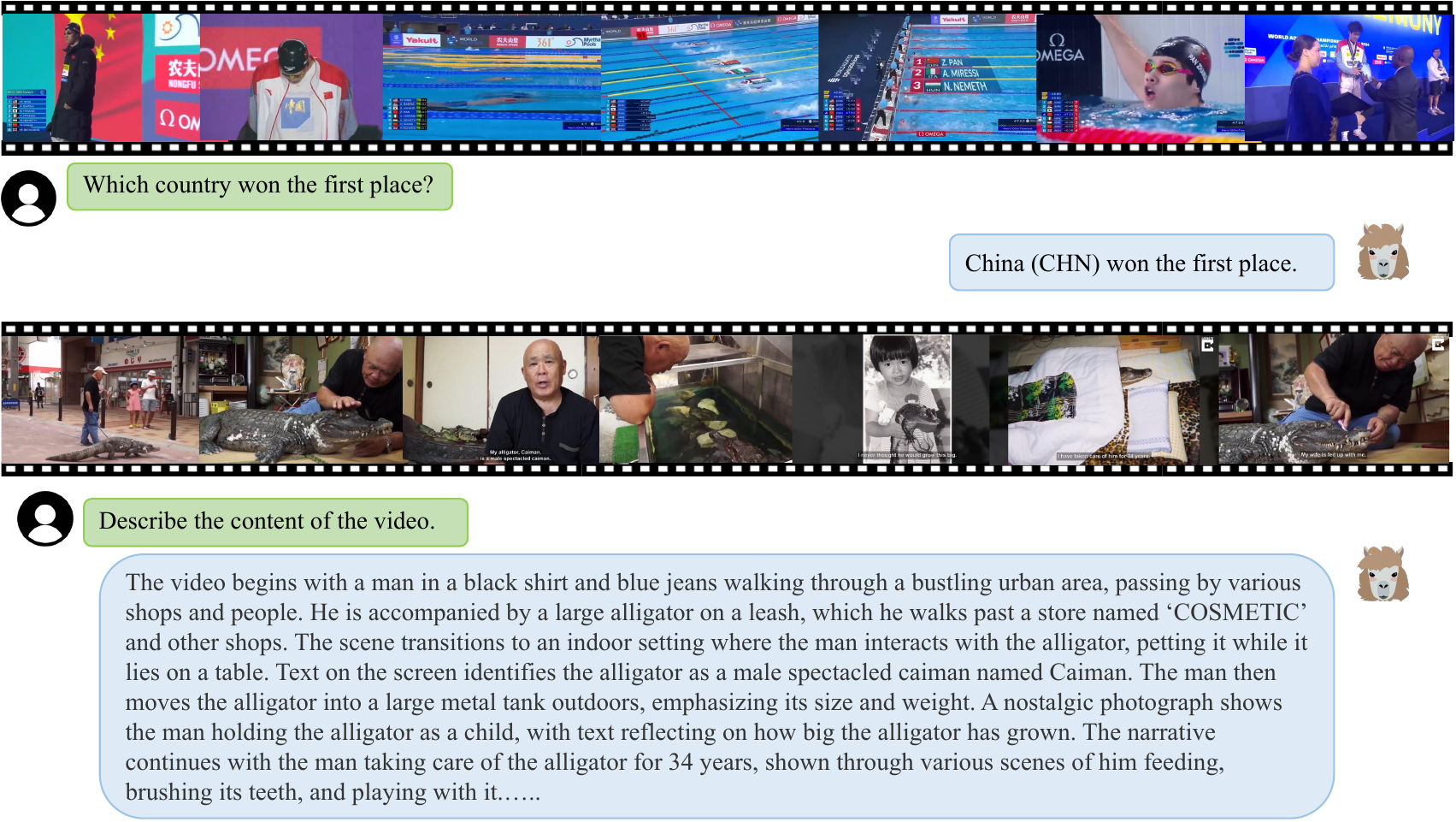}
    \vspace{3mm}
    \label{fig:image2}
  \end{subfigure}
   \begin{subfigure}[t]{\textwidth}
    \centering
    \includegraphics[width=0.9\textwidth]{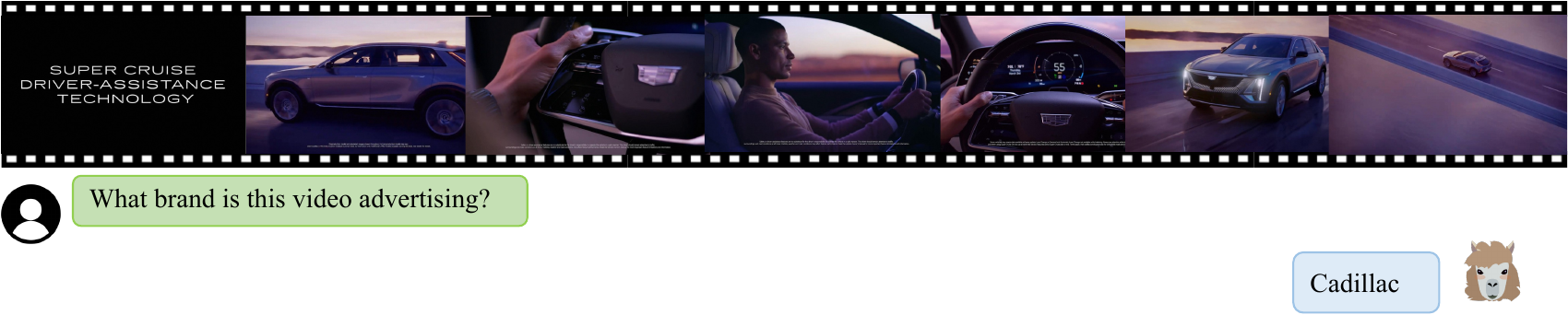}
    \label{fig:image3}
  \end{subfigure}
  \caption{Qualitative results on video QA and video caption task. For each video among these cases, we sample 96 frames.}
  \label{fig:overall}
\end{figure*}

\begin{table*}[ht]
  \centering
   \scalebox{1.2}{ 
    \begin{tabular}{c|ccccc}
      \toprule
      Dataset & Avg words &Avg duration&Total duration&Caption&Text \\
      \midrule
      HowTo100M \cite{howto100m} &4.0&3.6s&135Khr&136M&ASR\\ 
      LSMDC \cite{lsmdc}&7.0&4.8s&158h&118K&Manual\\
      DiDeMo \cite{didemo}&8.0&6.9s&87h&27K&Manual\\
      ActivityNet \cite{activitynet}&13.5&36.0s&849h&100K&Manual\\
      YouCook2 \cite{youcook2}&8.8&19.6s&176h&14K&Manual\\
      Panda-70M \cite{panda70m}&13.2&8.5s&167Khr&70.8M&Open-source Model\\
      MSR-VTT \cite{msrvtt}&9.3&15.0&40h&10K&Open-source Model\\
      VATEX \cite{vatex}&15.2&10s&115h&413K&Manual \\
      HD-VILA-100M \cite{hdvila100m}&17.6&11.7s&760Khr&103M&ASR\\
      InternVid \cite{internvid}&32.5&13.4s&372Khr&234M&Open-source Model \\
      ShareGPT4Video\cite{sharegpt4video}&300&27.0s&0.2Khr&40K&GPT-4V\\
      LLaVA-Video-178K \cite{llava-video}&-&40.4s&2Khr&178K&GPT-4O\\
      \midrule
      Ours &4967&666.9s&8Khr&40K&Manual\\
      \bottomrule
    \end{tabular}
    }
    \caption{Comparison of our dataset and others video-language datasets.}
  \label{tab:dataset}
\end{table*}


\end{document}